\documentclass{article}%
\usepackage{amsmath}
\usepackage{amsfonts}
\usepackage{amssymb}
\usepackage{graphicx}
\usepackage{algorithmic}
\usepackage{algorithm}%
\providecommand{\U}[1]{\protect\rule{.1in}{.1in}}

\begin{document}

\title{Multi-Attention Multiple Instance Learning}
\author{Andrei V. Konstantinov and Lev V. Utkin\\Peter the Great St.Petersburg Polytechnic University\\St.Petersburg, Russia\\e-mail: andrue.konst@gmail.com, lev.utkin@gmail.com}
\date{}
\maketitle

\begin{abstract}
A new multi-attention based method for solving the MIL problem (MAMIL), which
takes into account the neighboring patches or instances of each analyzed patch
in a bag, is proposed. In the method, one of the attention modules takes into
account adjacent patches or instances, several attention modules are used to
get a diverse feature representation of patches, and one attention module is
used to unite different feature representations to provide an accurate
classification of each patch (instance) and the whole bag. Due to MAMIL, a
combined representation of patches and their neighbors in the form of
embeddings of a small dimensionality for simple classification is realized.
Moreover, different types of patches are efficiently processed, and a diverse
feature representation of patches in a bag by using several attention modules
is implemented. A simple approach for explaining the classification
predictions of patches is proposed. Numerical experiments with various
datasets illustrate the proposed method.


\end{abstract}

\section{Introduction}

The Multiple Instance Learning (MIL) as a type of weakly supervised learning
is a framework which is applied to many applications, including the drug
activity prediction \cite{Dietterich-etal-97}, detecting the lung cancer
\cite{Zhu-etal-08}, the protein function annotation \cite{Wei-etal-19},
histology \cite{Hagele-etal-20,Laak-etal-21,Yamamoto-etal-2019}, and many
others. The corresponding survey works describing various MIL problem
statements and applications can be found in
\cite{Amores-13,Babenko-08,Carbonneau-etal-18,Cheplygina-etal-19,Quellec-etal-17,Yao-Zhu-etal-20,Zhou-04}%
.

In contrast to traditional machine learning models, MIL deals with a set of
bags that are labeled such that each bag consists of many unlabeled instances,
i.e. instances in the bags have no label information, and we have weakly
annotated data. MIL aims to train a classifier that assign labels to testing
bags or to assign labels to unlabeled instances in bags. The standard MIL
assumption states that all negative bags contain only negative instances, and
that positive bags contain at least one positive instance. It is important
that this is one of the possible MIL problem statements. Other statements are
when, for example, instances are annotated by an expert or when instances as
well as bags do not have labels \cite{Srinidhi-etal-21}.

Many MIL models which adapt various conventional models to solving MIL tasks
have been proposed in literature, for example, the KNN is adapted to the
citation-kNN \cite{Wang-Zucker-00}, the SVM to mi-SVM and MI-SVM
\cite{Andrews-etal-02}, decision trees to ID3-MI \cite{Chevaleyre-Zucker-01},
convolutional neural networks to the Multiple Instance Learning Convolutional
Neural Network \cite{Kraus-Ba-Frey-16,Sun-Han-etal-16,Wang-Yan-etal-18}.

One of the promising approaches to improve the MIL models is to use the
attention mechanism. An attention-based MIL was proposed in
\cite{Pappas-PopescuBelis-17} where the linear regression model was replaced
by a one-layer neural network with single output. Following this work, several
methods of MIL based on the attention mechanism have been developed, including
Deep Attention Multiple Instance Survival Learning \cite{Yao-Zhu-etal-20},
ProtoMIL \cite{Rymarczyk-etal-21}, MHAttnSurv \cite{Jiang-etal-21}, the
loss-attention MIL \cite{shi2020loss}, MILL \cite{tang2021mill}. There are
other MIL methods using the attention mechanism, which can be found in
\cite{Fuster-etal-21,Li-Li-Eliceiri-21,Wang-Zhou-20}. An interesting method
for applying attention to the MIL problem is provided by Ilse et al.
\cite{Ilse-etal-18}. We will use this method as a basic attention-based
approach to MIL and call it as AbDMIL (Attention-based Deep Multiple Instance
Learning) below.

However, the above methods have some disadvantages. First of all, they do not
take into account the neighboring patches or instances which may significantly
impact on a prediction especially when images are considered as bags. Our
experiments show that neighbors can be regarded as an additional information
about a patch. That is why we propose to analyze every patch jointly with $K$
neighbors where $K$ can be viewed as a tuning parameter. Moreover, we propose
to transform all neighbors to a single embedding which collects all
information from neighbors into one vector. This is implemented by means of
the attention. In order to simplify the analysis, we consider only adjacent
patches because they are the most informative. Another important advantage of
taking into account neighbors is that patches can be selected as small as
possible because each patch jointly with neighbors can cover an increased area
of the whole image. Smaller patches have simpler procedures for their
processing. The larger the size of a patch, the less the possibility of
interpreting the result (in the extreme case, the patch is the whole image),
and also the more difficult the task of training the model under condition of
small sizes of the training instances. In general case, if we assume that
instances of a bag impact on a bag label jointly with their neighbors, then
the classification problem could be reformulated in terms of the standard MIL
problem statement. For example, we could form new enlarged instances
consisting of the initial instances with neighbors. However, the order of
neighbors in this case will be taken into account. This property is not
desirable. Another difficulty is variability of neighbors number which may
take place in many applications, for example, in computational histopathology.
Moreover, if we apply a MIL algorithm to new instances in this case, then
identical parts of different new instances will be performed differently. In
order to overcome this disadvantage, we propose to apply an encoder network to
every instance and to compose joint embeddings of instances with neighbors.

Second, most MIL methods use the attention mechanism under condition that
attention weights allow us to find key instances in the whole image. This
application of the attention is very important because its use enhances the
classification quality. However, the attention can be used in a more wide
sense. We introduce the multiple attention mechanism which is based on
training a set of templates. The templates may be very useful when patches or
instances corresponding to a positive label are significantly different. Each
template and the corresponding attention is responsible for every type of
patches. The templates are viewed as trainable parameters which are trained
under condition that they should be different in order to take into account
different aspects of instances. Another peculiarity of the proposed scheme is
that we avoid the concatenation operation with embeddings obtained by means of
the attention modules. In other words, we propose to aggregate embeddings of
templates by using the attention mechanism instead of the direct use of a
neural network to a set of concatenated embeddings. An additional attention is
used to processing all these embeddings. As a result, we do not need to train
the encoder neural network on every new type of patches. It is trained once,
and only templates are trained for new patches.

In sum, we propose a multiple attention scheme for solving the MIL problem
called MAMIL, where one of the attention modules takes into account adjacent
patches or instances (Neighborhood attention), several attention modules
(Template Attentions) are used to get a diverse feature representation of each
patch, and one attention (Final Attention) is used to unite different feature
representations and to classify each patch (instance) and the whole bag.

Our contributions can be summarized as follows:

\begin{enumerate}
\item A new MIL model, which takes into account the neighboring patches or
instances of each analyzed patch.

\item A way to join all information available in neighbors of a patch in the
form of a united embedding by means of the neighborhood attention, which takes
into account the proximity of each neighbor to the corresponding patch.

\item A way to efficiently process different types of patches and to implement
diverse feature representation of patches in a bag by using several attention modules.

\item A combined representation of patches and their neighbors in the form of
an embedding of a small dimensionality for simple classification.

\item Explaining why a patch is classified in a certain way.
\end{enumerate}

Numerical experiments with various datasets illustrate the proposed method. In
particular, four datasets constructed from the well-known MNIST dataset
\cite{LeCun-etal-98} to get the specific MIL\ datasets are studied. Five
datasets Musk1, Musk2 \cite{Dietterich-etal-97}, Fox, Tiger, Elephant
\cite{Andrews-etal-02} with numerical features are used to perform tabular
data. Since one of the important applied areas, where MIL can be viewed as a
main and inherent tool, is the computational histopathology, then we study the
proposed method by using the Breast Cancer Cell Segmentation dataset
\cite{Gelasca-etal-08} consisting of histopathology images.

The paper is organized as follows. Related work can be found in Section 2. A
brief introduction to MIL and basics of the attention mechanism are given in
Section 3. The proposed method and the corresponding general scheme are
provided in Section 4. Computing the instance importance for the explanation
purposes is considered in Section 5. A brief discussion of introduced template
properties can be found in Section 6. Numerical experiments are provided in
Section 7. Concluding remarks can be found in Section 8.

\section{Related work}

\textbf{MIL}. Many models have been proposed to modify various machine
learning methods for solving the MIL problem, including the citation-kNN
\cite{Wang-Zucker-00}, the mi-SVM and MI-SVM \cite{Andrews-etal-02}, the
ID3-MI \cite{Chevaleyre-Zucker-01}. MIL models using ensemble-based
approaches, for example, AdaBoost, Random Forest, etc., were developed and
presented in \cite{Auer-Ortner-04,Leisner-etal-10,Mei-Zhu-14}. A comparison of
some ensemble-based MIL models can be found in \cite{Taser-etal-19}.

The large importance of MIL in many applications motivated developing models
based on neural networks and convolutional neural network
\cite{Doran-Ray-16,Kraus-Ba-Frey-16,Sun-Han-etal-16,Wang-Yan-etal-18}. We can
also point out such models as MI-BDNN \cite{Xu-16}, Deep MIML
\cite{Feng-Zhou-17}, MI-ELM \cite{Liu-Zhou-etal-16}.

Elements of the explanation methods
\cite{Arrieta-etal-2020,Guidotti-etal-21,Li-Xiong-etal-21,Rudin-etal-21} were
used to provide interpretability of the MIL models \cite{Rymarczyk-etal-21}.

A huge amount of methods and models were developed to apply MIL to the
computational histopathology where histology images can be viewed as bags and
are often represented as a set of small parts (patches, cells) which can be
referred to as \textquotedblleft instances\textquotedblright. Many survey
papers \cite{Hagele-etal-20,Laak-etal-21,Quellec-etal-17,Srinidhi-etal-21}
study various aspects of the computational histopathology as a MIL problem.

Various deep learning MIL models provide an important opportunity to improve
accuracy of the MIL predictions, to correctly classify bags as well as their
instances. However, a more interesting approach based on applying the
attention mechanism to solving the MIL\ problem opens another direction in
development of the accurate classification methods of MIL.

\textbf{MIL and attention. }In spite of a few papers devoted to application of
the attention mechanism \cite{Bahdanau-etal-14,Zhang2021dive} to solving the
MIL\ problem, it can be regarded as a promising approach to improve the MIL
models. Starting from the attention-based MIL \cite{Pappas-PopescuBelis-17}
and following this work, several interesting models using the attention
mechanism have been developed. They, for example, include DeepAttnMISL (Deep
Attention Multiple Instance Survival Learning) \cite{Yao-Zhu-etal-20},
MHAttnSurv (Multi-Head Attention for Survival Prediction) \cite{Jiang-etal-21}%
, ProtoMIL (Multiple Instance Learning with Prototypical Parts)
\cite{Rymarczyk-etal-21}, SA-AbMILP (Self-Attention Attention-based MIL
Pooling) \cite{Rymarczyk-etal-21a}, the loss-attention MIL (the instance
weights are calculated based on the loss function) \cite{shi2020loss}, DSMIL
(Dual-stream Multiple Instance Learning) \cite{Li-Li-Eliceiri-21} MILL
(Multiple Instance Learning--based Landslide classification)
\cite{tang2021mill}, AbDMIL \cite{Ilse-etal-18}. There are other MIL methods
using the attention mechanism, which can be found in
\cite{Fuster-etal-21,Qi-etal-17,Wang-Zhou-20}. The aforementioned methods
propose approaches to enhance the MIL classification quality by using the
attention. However, they do not take into account neighboring patches of each
analyzed patch and do not propose to aggregate information which can be
elicited from the neighborhood. In contrast to these methods, the proposed
MAMIL provides an attention-based way for aggregating and incorporating the
neighbors into the general MIL. MAMIL provides a flexible diverse
representation of a bag structure in the form of templates which allow us to
consider different aspects of instances and to take into account new types of
patches by training only attention models.

\section{Preliminary}

\subsection{Multiple Instance Learning}

According to the MIL problem statement, bags have class labels, but instances
are unlabeled. This is the weakly supervised learning problem. The lack of
labels for instances is a key peculiarity of MIL which motivates to solve two
tasks. The first one is to annotate instances of bags. The second task is to
classify new bags. In order to solve the tasks, we have to define conditions
which connect the instance labels and the bag classes. Let $X$ be a bag
defined as a set of feature vectors $X=\{\mathbf{x}_{1},...,\mathbf{x}_{m}\}$.
Each instance is represented as feature vector $\mathbf{x}_{i}$ in feature
space $\mathcal{X}$. It can be mapped to a class by some function
$f:\mathcal{X}\rightarrow\{0,1\}$, where the negative and positive classes
denoted as $y_{1},...,y_{m}$ correspond to 0 and 1, respectively. Classes
$y_{1},...,y_{m}$ remain unknown during training. It should point out that
binary classes is a special case of a general classification problem where the
number of classes may be arbitrary. The number of instances $m$ can also vary
for different bags. We will denote bags by capitals and instances by bold letters.

Let us define conditions of the class label assignment to bags. The most
common condition is that negative bags contain only negative instances, and
positive bags contain at least one positive instance \cite{Carbonneau-etal-18}%
. This implies that the bag classifier $g(X)$ is defined by%
\begin{equation}
g(X)=\left\{
\begin{array}
[c]{cc}%
1, & \exists\mathbf{x}\in X:f(\mathbf{x})=1,\\
0, & \text{otherwise.}%
\end{array}
\right.
\end{equation}

Another more general condition is when a threshold $\theta$ is introduced to
define the bag classifier $g(X)$, i.e., there holds%
\begin{equation}
g(X)=\left\{
\begin{array}
[c]{cc}%
1, & \theta\leq\sum_{\mathbf{x}\in X}f(\mathbf{x}),\\
0, & \text{otherwise.}%
\end{array}
\right.
\end{equation}

The first condition can be regarded as a special case of the second one when
$\theta=1$. Only the first condition is used below.

\subsection{Basics of the attention mechanism}

The attention mechanism can be regarded as a tool by which a neural network
can automatically distinguish the relative importance of features and weigh
the features for enhancing the classification accuracy. It can be viewed as a
learnable mask which emphasizes relevant information in a feature map. It is
pointed out in \cite{Chaudhari-etal-2019,Zhang2021dive} that the original idea
of attention can be understood from the statistics point of view applying the
Nadaraya-Watson kernel regression model \cite{Nadaraya-1964,Watson-1964}.
Given $n$ training instances $S=\{(\mathbf{x}_{1},y_{1}),(\mathbf{x}_{2}%
,y_{2}),...,(\mathbf{x}_{n},y_{n})\}$, in which $\mathbf{x}_{i}=(x_{i1}%
,...,x_{id})\in\mathbb{R}^{d}$ represents a feature vector involving $d$
features and $y_{i}\in\mathbb{R}$ represents the regression outputs, the task
of regression is to construct a regressor $f:\mathbb{R}^{d}\rightarrow
\mathbb{R}$ which can predict the output value $y$ of a new observation
$\mathbf{x}$, using available training data $S$. The similar task can be
formulated for the classification problem.

The original idea behind the attention mechanism is to replace the simple
average of outputs $y^{\ast}=n^{-1}\sum_{i=1}^{n}y_{i}$ for estimating the
regression output $y$ corresponding to a new input feature vector $\mathbf{x}$
with the weighted average in the form of the Nadaraya-Watson regression model
\cite{Nadaraya-1964,Watson-1964}:%
\begin{equation}
y^{\ast}=\sum_{i=1}^{n}\alpha(\mathbf{x},\mathbf{x}_{i})y_{i},
\end{equation}
where weight $\alpha(\mathbf{x},\mathbf{x}_{i})$ conforms with relevance of
the $i$-th training instance to the vector $\mathbf{x}$.

In other words, according to the Nadaraya-Watson regression model, to estimate
the output $y$ of an input variable $\mathbf{x}$, training outputs $y_{i}$
given from a dataset weigh in agreement with the corresponding input
$\mathbf{x}_{i}$ locations relative to the input variable $\mathbf{x}$. The
closer an input $\mathbf{x}_{i}$ to the given variable $\mathbf{x}$, the
greater the weight assigned to the output corresponding to $\mathbf{x}_{i}$.

One of the original forms of weights is defined by a kernel $K$ (the
Nadaraya-Watson kernel regression \cite{Nadaraya-1964,Watson-1964}), which can
be regarded as a scoring function estimating how vector $\mathbf{x}_{i}$ is
close to vector $\mathbf{x}$. The weight is written as follows
\cite{Zhang2021dive}:
\begin{equation}
\alpha(\mathbf{x},\mathbf{x}_{i})=\frac{K(\mathbf{x},\mathbf{x}_{i})}%
{\sum_{j=1}^{n}K(\mathbf{x},\mathbf{x}_{j})}.
\end{equation}

The above expression is an example of weights in nonparametric attention
\cite{Zhang2021dive}. In terms of the attention mechanism
\cite{Bahdanau-etal-14}, vector $\mathbf{x}$, vectors $\mathbf{x}_{i}$ and
outputs $y_{i}$ are called as the query, keys and values, respectively. Weight
$\alpha(\mathbf{x},\mathbf{x}_{i})$ is called as the attention weight.
Therefore, the standard attention applications are often represented in terms
of queries, keys and values, and the attention weights are expressed through
these terms.

Generally, weights $\alpha(\mathbf{x},\mathbf{x}_{i})$ can be extended by
incorporating learnable parameters. For example, if we denote $\mathbf{q=W}%
_{q}\mathbf{x}$ and $\mathbf{k}_{i}\mathbf{=W}_{k}\mathbf{x}_{i}$ referred to
as the query and key embeddings, respectively, then the attention weight can
be represented as:%

\begin{equation}
\alpha(\mathbf{x},\mathbf{x}_{i})=\text{\textrm{softmax}}\left(
\mathbf{q}^{\mathrm{T}}\mathbf{k}_{i}\right)  =\frac{\exp\left(
\mathbf{q}^{\mathrm{T}}\mathbf{k}_{i}\right)  }{\sum_{j=1}^{n}\exp\left(
\mathbf{q}^{\mathrm{T}}\mathbf{k}_{j}\right)  }, \label{Expl_At_12}%
\end{equation}
where $\mathbf{W}_{q}$ and $\mathbf{W}_{k}$ are matrices of parameters which
are learned, for example, by incorporating an additional feed forward neural
network within the system architecture.

Several attention weights defining the attention modules have been proposed
for different applications. They can be divided into the additive attention
\cite{Bahdanau-etal-14} and multiplicative or dot-product attention
\cite{Luong-etal-2015,Vaswani-etal-17}. As pointed out by
\cite{Niu-Zhong-Yu-21}, the attention modules can be also classified as
general attention, concat attention, and a location-based attention modules
\cite{Luong-etal-2015}. In particular, the general attention uses learnable
parameters for keys and queries as it is illustrated in \ref{Expl_At_12}
(parameters $\mathbf{W}_{q}$ and $\mathbf{W}_{k}$). The concat attention uses
the concatenation of keys and queries. In the location-based attention, the
scoring function depends only on queries and does not depend on keys. A list
of common attention types can be found in \cite{Niu-Zhong-Yu-21}.

\section{The proposed method}

Suppose that there are $N$ bags $X_{1},...,X_{N}$ with labels $Y_{1}%
,...,Y_{N}$. Every bag consists of $m_{k}$ patches or instances, $k=1,...,N$.

\begin{figure}
[ptb]
\begin{center}
\includegraphics[
height=2.2459in,
width=6.9764in
]%
{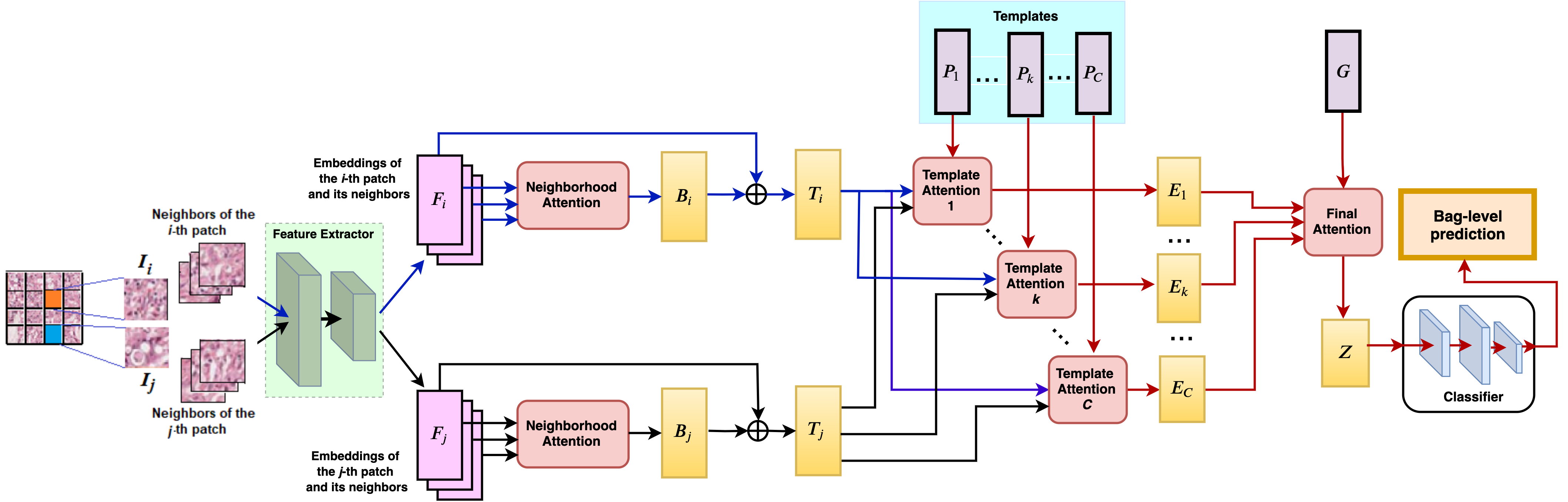}%
\caption{A scheme illustrating two paths of the patch processing in the neural
network which implement the proposed algorithm}%
\label{f:main_scheme}%
\end{center}
\end{figure}

Let us consider the main algorithm of using neighbors of each patch and
learnable templates which can be regarded as a result of applying attention
modules. A scheme of the algorithm is shown in Fig. \ref{f:main_scheme}.%

Suppose we have $N$ bags in the form of histological images. It should be
noted that the histological application is taken to make a more clear
illustration of the proposed algorithm. At the same time, we will study
datasets different from the histological application in the same way by using
this algorithm. Let us consider an image or a bag with the number of patches
$m$. It is shown in Fig. \ref{f:main_scheme} as the histological image. We
consider the $i$-th and the $j$-th patches of the image which are highlighted
in different colors.

All patches are fed to the input of a trained neural network (encoder or
feature extractor) for computing their embedding $F_{i}$, i.e.,%

\begin{equation}
F_{i}=Conv(I_{i}),\ i=1,...,m.
\end{equation}
Embeddings $F_{i}$ are produced to reduce the patch dimensionality.

For every patch, we also select a set of neighbors of adjacent patches. Let
$(x_{i},y_{i})$ and $\mathcal{M}=\{1,...,m\}$ be integer coordinates of the
$i$-th patch in the whole image and the index set of all patches in the whole
image, respectively. Then the index set $\mathcal{N}_{i}\subset\mathcal{M}$ of
neighbors of the $i$-th patch is determined as
\begin{equation}
\mathcal{N}_{i}=\{j\in\mathcal{M}~|~0<\max(|x_{i}-x_{j}|,|y_{i}-y_{j}|)\leq
d\}.
\end{equation}

In our experiments the parameter $d$ is taken equal to $1$, but it can
generally be arbitrary and depends on a size of patches. Moreover, other
definitions of neighbors can be applied, for example, adjacent patches. The
proposed algorithm depicted in Fig. \ref{f:main_scheme} is general, and it is
not restricted by a certain definition of neighbors. The idea to use the
neighbor patches is based on the assumption that these patches may carry some
useful information about a structure of the image. Moreover, the use of
neighbors can be viewed as some kind of augmentation which may improve the
classification results. Therefore, the next questions are how to take into
account the patch neighbor information, how to perform identical neighbors of
different patches, how to perform different numbers of neighbors. The answer
is to use the attention mechanism. Namely, the joint embedding $B_{i}$ of its
neighbors with numbers from the index set $\mathcal{N}_{i}$ of neighbors of
the $i$-th patch is computed by means of the attention mechanism.

The embedding $B_{i}$ of neighbors is determined as%
\begin{equation}
B_{i}=\sum_{j\in\mathcal{N}_{i}}\alpha_{j}^{(i)}F_{j}. \label{Atten_MIL_20}%
\end{equation}

Here weight $\alpha_{j}^{(i)}$ of the $j$-th neighbor is determined in
accordance with relevance of the $j$-th patch embedding to vector $F_{i}$,
i.e. it is determined in accordance with proximity of embedding $F_{j}$,
$j\in\mathcal{N}_{i}$, of the corresponding neighbor patch to embedding
$F_{i}$ of the analyzed patch $I_{i}$. Hence, we use the original attention
weights computed as follows:
\begin{equation}
\alpha_{j}^{(i)}=\mathrm{softmax}\left(  \mathbf{s}^{(i)}\right)  =\frac
{\exp\left(  s_{j}^{(i)}\right)  }{\sum_{t}\exp\left(  s_{t}^{(i)}\right)  },
\label{Atten_MIL_21}%
\end{equation}
where $s_{j}^{(i)}$ is the attention scoring function that maps two vectors
$F_{i}$ and $F_{j}$ to a scalar, i.e.,
\begin{equation}
s_{j}^{(i)}=\mathrm{score}_{nb}\left(  F_{i},F_{j}\right)  .
\end{equation}

It is defined in the proposed algorithm as%
\begin{equation}
\mathrm{score}_{nb}\left(  F_{i},F_{j}\right)  =F_{i}^{\mathrm{T}}\tanh\left(
V_{nb}F_{j}\right)  ,
\end{equation}
where $V$ is the matrix of learnable parameters.

In sum, we get a single embedding $B_{i}$ for each $F_{i}$, which contains
information about neighbors. It allows us to get joint embedding of the $i$-th
patch embedding $F_{i}$ and the embedding of neighbors. This can be done by
means of concatenation of embeddings, i.e., the embedding of the $i$-th patch
is concatenated with the joint embedding of neighbors $B_{i}$ to get the patch
embedding $T_{i}$ with neighborhood as:
\begin{equation}
T_{i}=\left(  F_{i},B_{j}\right)  .
\end{equation}

The concatenation operation is depicted in Fig. \ref{f:main_scheme} by symbol
$\oplus$. In fact, each bag is represented by the set of embeddings $T_{i}$ now.

Another important concept proposed in the algorithm is a set of embedding
templates which are regarded as learnable vectors. Moreover, the number of
templates is the hyperparameter $C$ whose value determines the quality of
classification. Templates can be regarded as implementation of multiple
attention applied to the set of embeddings $T_{i}$. They are different due to
different initialization rules of the neural network. 

Let us denote the set of templates as $\mathcal{P}=\{P_{1},...,P_{C}\}$ and
the $k$-th template as $P_{k}\in\mathcal{P}$. Then every template attention
produces the corresponding aggregate embedding $E_{k}$ which is computed as
follows:%
\begin{equation}
E_{k}=\sum_{i=1}^{m}\beta_{i}^{(k)}T_{i},\ k=1,...,C,
\end{equation}
where
\begin{equation}
\beta_{i}^{(k)}=\mathrm{softmax}\left(  \left\langle \mathrm{score}\left(
P_{k},T_{i}\right)  \right\rangle _{i}\right)  .
\end{equation}

It can be seen from the above that the $k$-th aggregate embedding $E_{k}$ is
the weighted mean of all patch embeddings with neighborhood from an image with
the weights defined by the $k$-th template. In other words, the $k$-th
aggregate embedding characterizes the whole image or the bag with respect to
the $k$-th template. Information about all patches and their neighbors is
contained in $C$ vectors $E_{1},...,E_{C}$ now.

Aggregate embeddings $E_{k}$ corresponding to all $C$ templates are grouped
into the overall vector $Z$ which corresponds to the whole image or the bag
and all templates:%
\begin{equation}
Z=\sum_{k=1}^{C}\gamma_{k}E_{k}, \label{Atten_MIL_30}%
\end{equation}
where
\begin{equation}
\gamma_{k}=\mathrm{softmax}\left(  \left\langle \mathrm{score}_{fin}\left(
G,E_{k}\right)  \right\rangle _{k}\right)  , \label{Atten_MIL_31}%
\end{equation}
$G$ is the global template as a training vector.

Template $G$ and the corresponding final attention determine which aggregate
embeddings $E_{k}$ and templates $P_{k}$ are important. The overall vector $Z$
is a feature representation of the whole bag or the image, and it takes into
account all peculiarities of the bag. It should be noted that we could
aggregate $E_{1},...,E_{C}$ by using their concatenation. However, the
aggregation of embeddings $E_{k}$ by means of the final attention mechanism
instead of their simple concatenation allows us to reduce the input dimension
of the classifier, thereby increasing the generalization ability of the model
and allowing possible training on small datasets. Moreover, the use of
attention mechanism before the classifier leads to model interpretability,
even if classifier is not interpretable.

It is proposed to use the classification layers $g_{\theta}(Z)$ with
parameters $\theta$, which defines the probability of class 1 at the bag level
(the whole image level). In the simplest case, the layers are linear with the
softmax, but an arbitrary classification neural network can also be used. It
is important that predictions can be interpreted even by using a complex
neural network.

We have described one path of the $i$-th patch processing. Fig.
\ref{f:main_scheme} illustrates two paths corresponding to processing of the
$i$-th and the $j$-th patches. However, the real implementation supposes to
process all patches of a bag in the same way.

The whole neural network is trained end-to-end using the stochastic gradient
descent and the Adam optimizer. At each step, an image (a bag) is selected
from the training set, then it is divided into patches. The patches are used
as input of the neural network. The result of using the neural network is an
estimate of the probability of a class which means, for example, whether the
image contains malignant cells or not. The loss function is determined based
on the obtained estimate and the whole image label. To update the neural
network weights, values of partial derivatives of the loss function are
determined for each trained parameter using an automatic differentiation
algorithm. Then the training parameter values are updated in a standard way.

Since templates play a role of different feature representations of each bag,
then the corresponding vectors should be different. This implies that the
standard loss function of the classification training is supplemented by the
additional term which provides the difference of vectors $P_{k}$, $k=1,...,C$.
Then the loss function is of the form:%
\begin{equation}
L=\frac{1}{N}\sum_{k=1}^{N}BCE(Y_{k},f(X_{k}))+\frac{2}{C(C-1)}\sum
_{i<j}(P_{i}^{T}\cdot P_{j})^{2}. \label{Atten_MIL_56}%
\end{equation}

Here the first term is the standard binary cross-entropy (BCE) loss function;
$f(X_{k})$ is the output of the whole neural network. The second term in the
loss function tries to make pairs of templates $P_{i}$ and $P_{j}$ as
different as possible.

\section{Computing the patch importance}

Another interesting advantage of the proposed model and the implementing
algorithm can be referred to the prediction interpretation or explanation. The
problem is to explain which patches have the largest impact on the prediction
of the bag class. It turns out that the proposed model provides the
opportunity to answer this question due to specific structure of the model,
namely, due to a connection between overall vector $Z$ and the patch
embeddings $T_{i}$, $i=1,...,m$.

The overall embedding $Z$ can be expressed through the weighted sum of
embeddings corresponding to patches. This follows from the following:
\begin{align}
Z  &  =\sum_{k=1}^{C}\gamma_{k}E_{k}=\sum_{k=1}^{C}\gamma_{k}\left(
\sum_{i=1}^{m}\beta_{i}^{(k)}T_{i}\right) \nonumber\\
&  =\sum_{i=1}^{m}\left(  \sum_{k=1}^{C}\gamma_{k}\beta_{i}^{(k)}\right)
T_{i}=\sum_{i=1}^{m}w_{i}T_{i}. \label{Atten_MIL_61}%
\end{align}

Here weight $w_{i}=\sum_{k=1}^{C}\gamma_{k}\beta_{i}^{(k)}$ can be regarded as
the \textit{importance} of the $i$-th patch with its neighborhood. It is easy
to show that there holds $\sum_{i=1}^{m}w_{i}=1$.

It should be noted that the importance of the $i$-th patch neighbors is
determined as $\alpha_{j}^{(i)}$ (see (\ref{Atten_MIL_20}) and
(\ref{Atten_MIL_21})). Hence, we can state the question how to connect the
\textit{importance} of the embedding $T_{i}$ taking into account neighbors
with the importance of patch $F_{i}$ itself. If we assume that all elements of
vector $T_{i}$ are equivalent, then importance measures of its two parts are
equal. However, $B_{i}$ may depend on $T_{j}$ if patches with indices $i$ and
$j$ are neighbor. This implies that the final importance $v_{i}$ of the $i$-th
patch is determined as%
\begin{equation}
v_{i}=w_{i}+\sum_{j\in N_{i}}\alpha_{j}^{(i)}w_{j}. \label{Atten_MIL_62}%
\end{equation}

\section{Again about templates}

The proposed templates have several advantages which are discussed below.

First of all, templates can be used for classification of types of instances.
This can be done by means of determining weights of the templates $\gamma_{k}$
for a particular image (see (\ref{Atten_MIL_30}) and (\ref{Atten_MIL_31})).
Having a small dataset with labels in the form of types of instances (not
binary), we can train a simple classifier that matches the vector $\left(
\gamma_{1},...,\gamma_{C}\right)  $ with the type of instances.

Another advantage is a simple possibility of adding new templates or excluding
unnecessary templates. Since the number of templates can be different without
changing the network weights, templates can be added or excluded after
training. Adding new templates corresponds to a scenario of appearance of a
new type of cells (the encoder has to be able to construct informative
embedding in this case). A new trainable vector of the template is added, and
all other network parameters are not trained and remain without changes. The
exclusion of templates can be carried out based on the statistics of template
weights over the entire dataset. Templates with the lowest total weight can be
excluded without reducing the prediction quality.

Templates can be visualized. For each template $P_{k}$, we can calculate its
proximity to all patch embeddings $T_{i}$ with neighborhoods of all dataset
images and find the patch with the largest value of $\mathrm{score}\left(
P_{k},T_{i}\right)  $. This patch can be regarded as an approximate
visualization of the template. An autoencoder reconstructing a patch image
from its embedding can be trained such that the patch image reconstructed from
a template can be viewed as the visualization of the template.

\section{Numerical experiments}

\subsection{Datasets for experiments}

We consider several dataset in order to study the proposed model.

First, we study the proposed model by applying the MNIST dataset which is a
commonly used large dataset of 28x28 pixel handwritten digit images
\cite{LeCun-etal-98}. It has a training set of 60,000 instances, and a test
set of 10,000 instances. The digits are size-normalized and centered in a
fixed-size image. The dataset is available at http://yann.lecun.com/exdb/mnist/.

To study how neighbors and templates impact on predictions, we perform four
datasets based on MNIST.

The first dataset called MNIST-MIL consists of groups (bags) of digits from
MNIST of random size from 6 to 12 digits in a bag such that the bag label is 1
if there is at least one digit \textquotedblleft9\textquotedblright\ among all
digits in the bag, otherwise label 0 is assigned to the bag.

The second dataset called MNIST-MIL-1 consists of groups (bags) of digits from
MNIST, which are composed similarly to the previous MNIST dataset, but label 1
is assigned to a bag if there is at least one pair of neighboring digits
\textquotedblleft9\textquotedblright\ and \textquotedblleft3\textquotedblright%
\ in the arbitrary order, otherwise label 0 is assigned to the bag.

In the third dataset called MNIST-MIL-2, bags are formed in the same way as in
MNIST-MIL-1, but label 1 is assigned to a bag if there is at least one digit
\textquotedblleft9\textquotedblright, which does not have neighbor
\textquotedblleft3\textquotedblright\ (on the left or on the right), otherwise
label 0 is assigned to the bag.

The fourth dataset called MNIST-MIL-3 consists of groups (bags) of digits from
MNIST, which are composed similarly to MNIST-MIL-2, but label 1 is assigned to
a bag if there is at least one digit \textquotedblleft9\textquotedblright,
which does not have neighbor \textquotedblleft3\textquotedblright, and there
is at least one digit \textquotedblleft7\textquotedblright, which does not
have neighbor \textquotedblleft4\textquotedblright, otherwise label 0 is
assigned to the bag.

We also use datasets Musk1, Musk2 (drug activity) \cite{Dietterich-etal-97},
Fox, Tiger, Elephant \cite{Andrews-etal-02} with numerical features to study
how the proposed method performs tabular data.

The Musk1 dataset contains 92 bags consisting of 476 instances with 166
features. The average bag size is $5.17$. The Musk2 dataset contains 102 bags
consisting of 6598 instances with 166 features. The average bag size is
$64.69$. Each dataset (Fox, Tiger and Elephant) contains exactly 200 bags
consisting of instances with 230 features. Numbers of instances in datasets
Fox, Tiger and Elephant are 1302, 1220 and 1391, respectively. The average bag
sizes of the datasets are $6.60$, $6.96$ and $6.10$, respectively.

In order to study the histology application of the proposed model and to
compare this model with results obtained in \cite{Ilse-etal-18}, we use the
Breast Cancer Cell Segmentation dataset \cite{Gelasca-etal-08} which consists
of 58 histopathology images with expert annotations. Images are used in breast
cancer cell detection with associated ground truth data available. The dataset
aims to validate methods for cell segmentation and their classification. The
dataset can be downloaded from https://www.kaggle.com/andrewmvd/breast-cancer-cell-segmentation.

Each image from the Breast Cancer Cell Segmentation dataset has the size
$896\times768$ pixels and is divided into $672$ non-intersecting patches of
size $32\times32$. Patches consisting of larger than $75\%$ of white pixels
are excluded from the dataset.

\subsection{MNIST-MIL results}

First, we study the proposed model trained on the dataset MNIST-MIL-1.
Randomly selected bags are grouped in accordance with their class and the
predicted class of the bag. The class of a bag in MNIST-MIL-1 is defined by
existence of neighboring digits \textquotedblleft9\textquotedblright\ and
\textquotedblleft3\textquotedblright. In order to correctly compare our
experiments with AbDMIL \cite{Ilse-etal-18}, we use the same model (LeNet5
\cite{LeCun-etal-98}) for training encoder depicted in Fig.
\ref{f:main_scheme} as \textquotedblleft Feature Extractor\textquotedblright%
\ whose output is $F_{i}$ as it is used in \cite{Ilse-etal-18}. Parameters of
the neural network and its training are shown in Tables 8-10 (see Appendix of
\cite{Ilse-etal-18}). In addition to the above, we use $C=10$ templates for
producing aggregate embeddings $E_{k}$. All embeddings $F_{i}$ and $B_{i}$
have the dimensionality which is equal to 128. Embeddings $T_{i}$, templates
$P_{i}$, aggregate embeddings $E_{k}$, vectors $G$ and $Z$ are of size $256$.

\begin{figure}
[ptb]
\begin{center}
\includegraphics[
height=2.7971in,
width=3.7565in
]%
{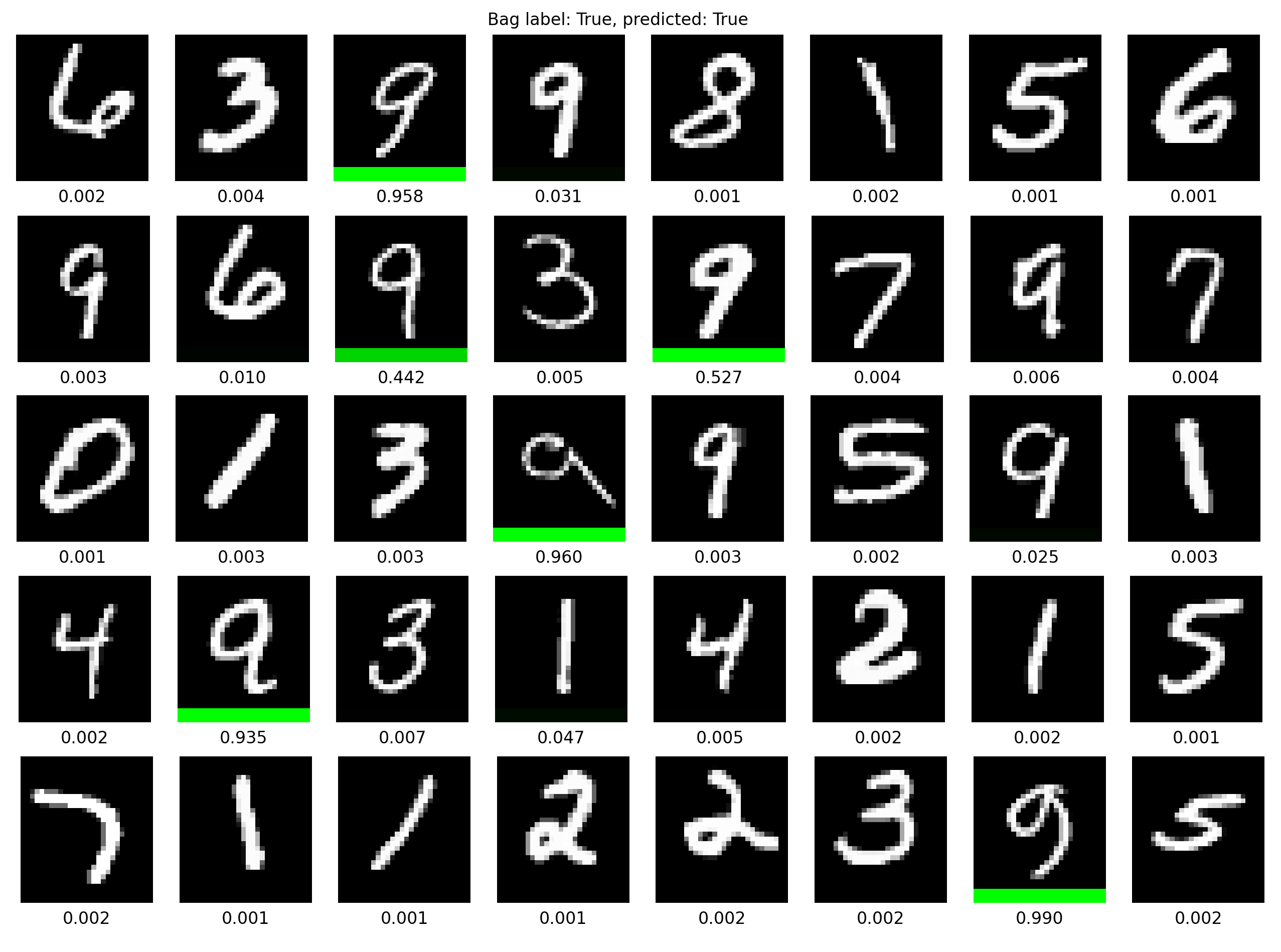}%
\caption{Examples of testing bags from MNIST-MIL-1, which belong to the true
positive subset}%
\label{f:tp_mnis}%
\end{center}
\end{figure}

The first subset of five randomly selected testing bags is shown in Fig.
\ref{f:tp_mnis}. Every bag belongs to class 1 (true) because it has
neighboring digits \textquotedblleft9\textquotedblright\ and \textquotedblleft%
3\textquotedblright. We take digits \textquotedblleft9\textquotedblright\ and
\textquotedblleft3\textquotedblright\ because they can be mistaken with many
digits, for example, with \textquotedblleft5\textquotedblright,
\textquotedblleft4\textquotedblright, \textquotedblleft2\textquotedblright,
\textquotedblleft8\textquotedblright. The algorithm successfully recognizes
bags as elements of class 1 because it finds digit \textquotedblleft%
9\textquotedblright\ with neighbor \textquotedblleft3\textquotedblright\ in
every bag. The importance $v_{i}$ of patches (digits) are shown under the
patches. It can be seen from Fig. \ref{f:tp_mnis} that digit \textquotedblleft%
9\textquotedblright\ has the largest value of $v_{i}$.%

\begin{figure}
[ptb]
\begin{center}
\includegraphics[
height=2.8124in,
width=3.7446in
]%
{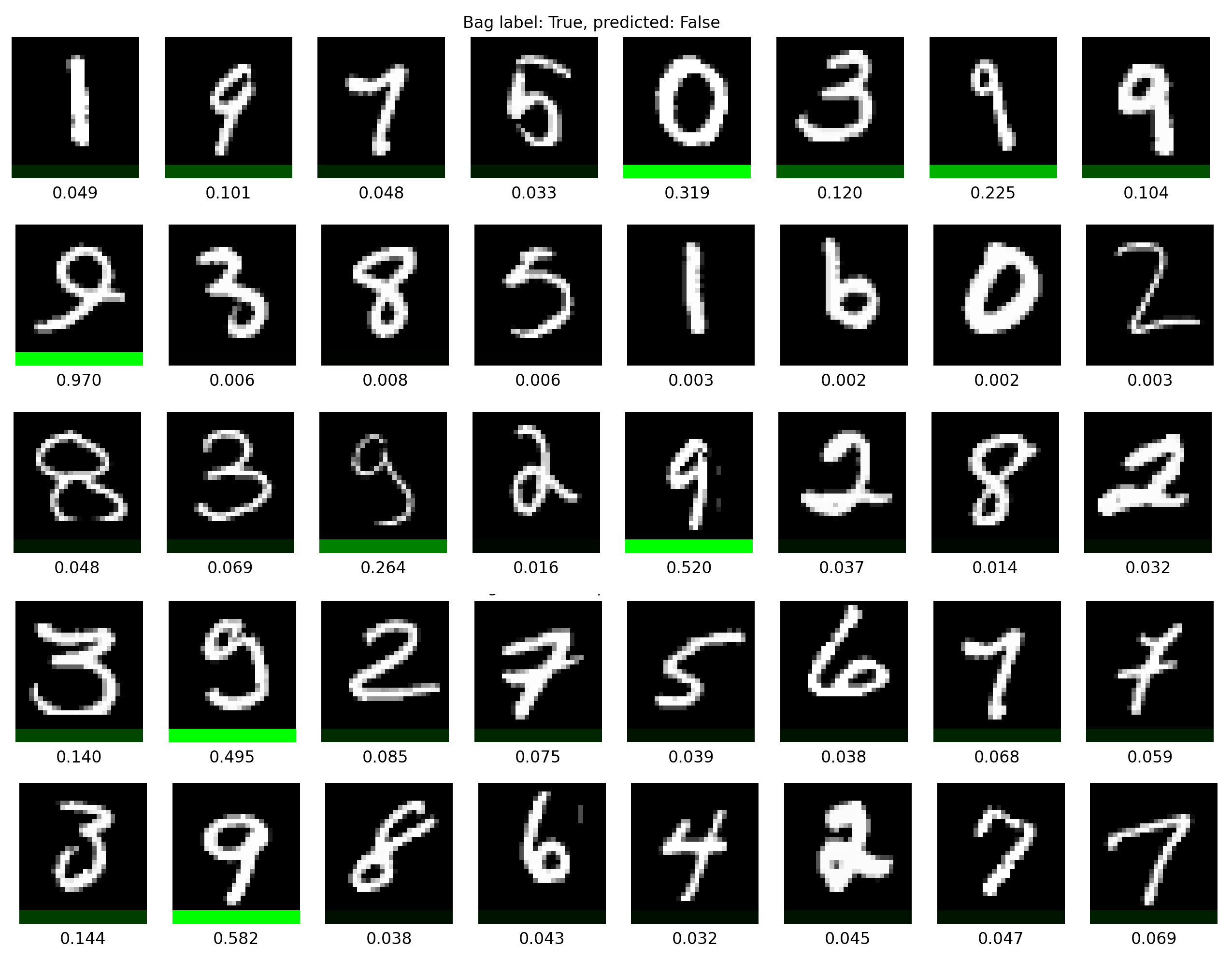}%
\caption{Examples of testing bags from MNIST-MIL-1, which belong to the false
negative subset}%
\label{f:tn01_mnis}%
\end{center}
\end{figure}

Another case is depicted in Fig. \ref{f:tn01_mnis} where examples of testing
bags, which belong to the false negative subset, are shown. One can see from
\ref{f:tn01_mnis} that every bag contains a pair of adjacent digits
\textquotedblleft9\textquotedblright\ and \textquotedblleft3\textquotedblright%
\ and can be regarded as belonging to class 1, but the classification model
incorrectly classifies digits \textquotedblleft9\textquotedblright\ as
different digits or as \textquotedblleft9\textquotedblright\ but with a
adjacent digit different from \textquotedblleft3\textquotedblright.%

\begin{figure}
[ptb]
\begin{center}
\includegraphics[
height=2.7773in,
width=3.7484in
]%
{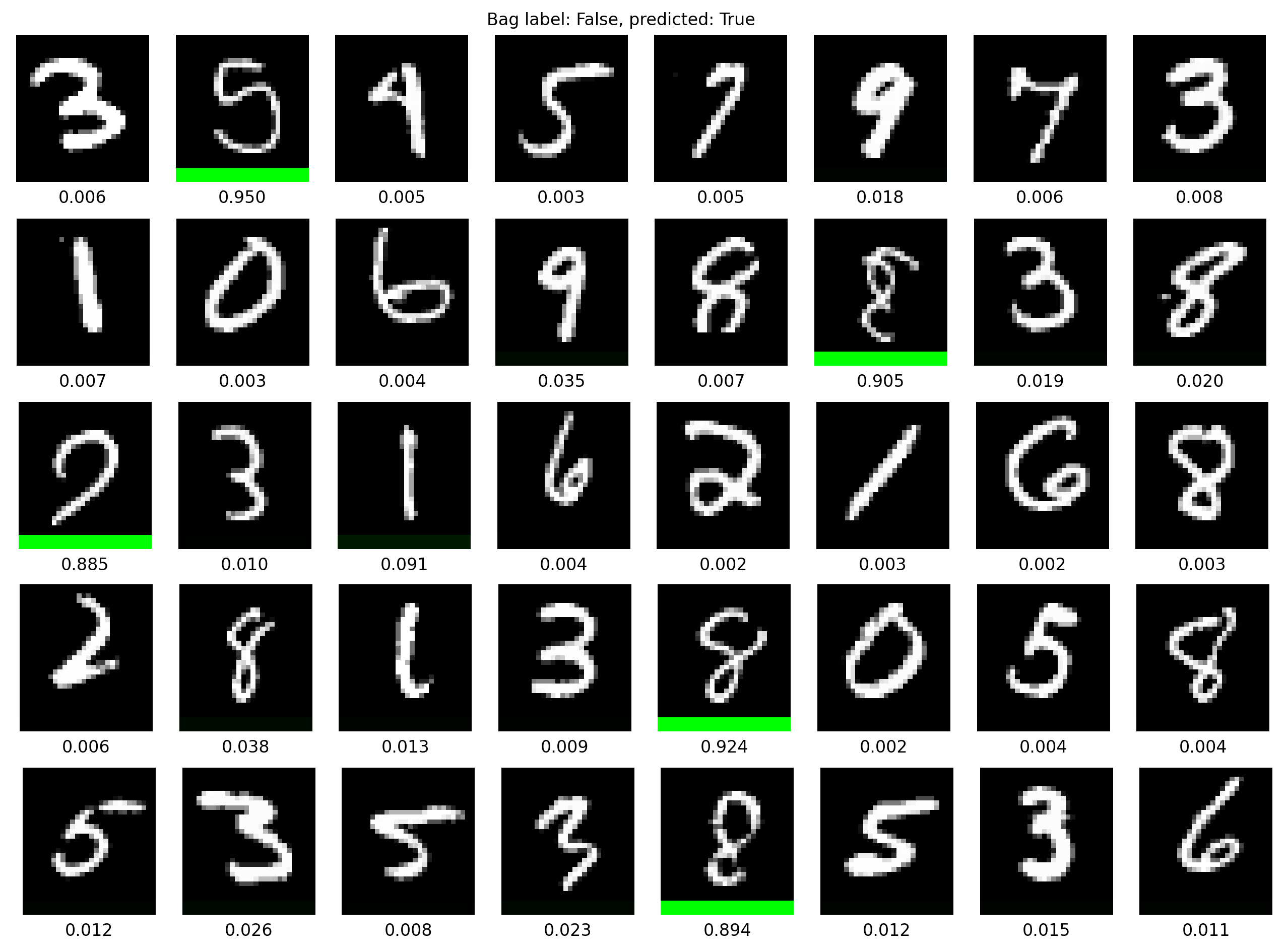}%
\caption{Examples of testing bags from MNIST-MIL-1, which belong to the false
positive subset}%
\label{f:fp01_mnis}%
\end{center}
\end{figure}

Fig. \ref{f:fp01_mnis} illustrates cases of false positive examples when bags
belong to class 0 because there are no adjacent pairs of digits
\textquotedblleft9\textquotedblright\ and \textquotedblleft3\textquotedblright%
. It is interesting to note that the model recognizes digits akin to
\textquotedblleft9\textquotedblright\ as \textquotedblleft9\textquotedblright%
\ with adjacent \textquotedblleft3\textquotedblright. For example, digit
\textquotedblleft5\textquotedblright\ in the first bag is akin to
\textquotedblleft9\textquotedblright, and there is digit \textquotedblleft%
3\textquotedblright\ on the left. However, this is an incorrect prediction. In
other words, the model correctly looks for the pair \textquotedblleft%
9\textquotedblright\ and \textquotedblleft3\textquotedblright\ and fails due
to the incorrect classification of \textquotedblleft9\textquotedblright.%

\begin{figure}
[ptb]
\begin{center}
\includegraphics[
height=2.9136in,
width=3.7334in
]%
{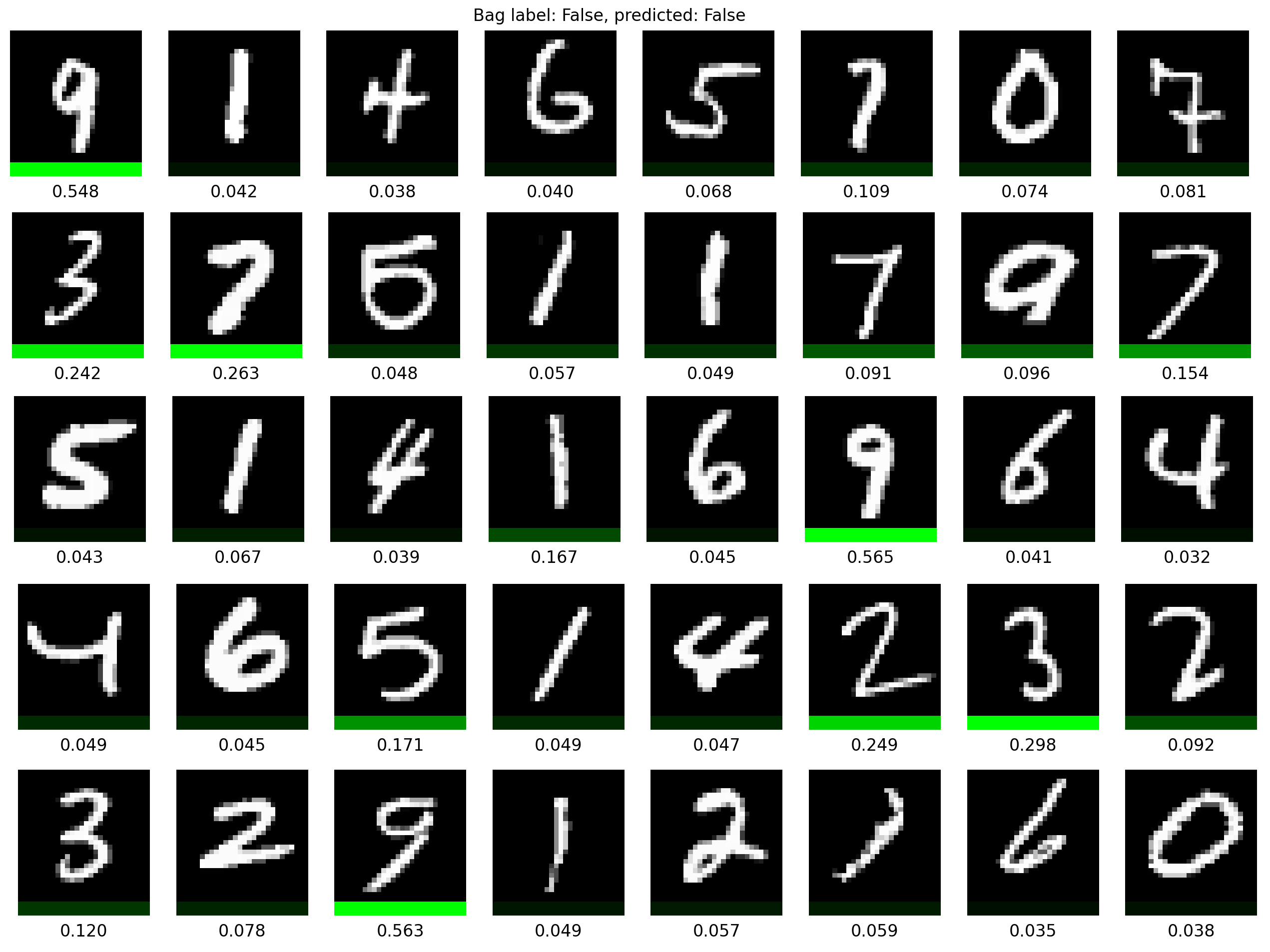}%
\caption{Examples of testing bags from MNIST-MIL-1, which belong to the true
negative subset}%
\label{f:fn01_mnis}%
\end{center}
\end{figure}

The last case of classification based on MNIST-MIL-1 is shown in Fig.
\ref{f:fn01_mnis} where bags belong to class 0 because there are no adjacent
digits \textquotedblleft9\textquotedblright\ and \textquotedblleft%
3\textquotedblright, and the model successfully classifies the bags as
negative. One can see from Fig. \ref{f:fn01_mnis} that the model correctly
recognizes \textquotedblleft9\textquotedblright\ in all cases except for the
second bag, but the recognized digits do not have adjacent \textquotedblleft%
3\textquotedblright. In the second bag, the model has detected the adjacent
pair \textquotedblleft3\textquotedblright\ and \textquotedblleft%
9\textquotedblright, but it incorrectly classifies the second digit as
\textquotedblleft9\textquotedblright.%

Let us consider the proposed model trained on the dataset MNIST-MIL-2.
Randomly selected bags are grouped in accordance with their class and the
predicted class of the bag. The class of a bag in MNIST-MIL-2 is defined by
existence at least one digit \textquotedblleft9\textquotedblright\ without
adjacent digit \textquotedblleft3\textquotedblright.

\begin{figure}
[ptb]
\begin{center}
\includegraphics[
height=2.7899in,
width=3.7034in
]%
{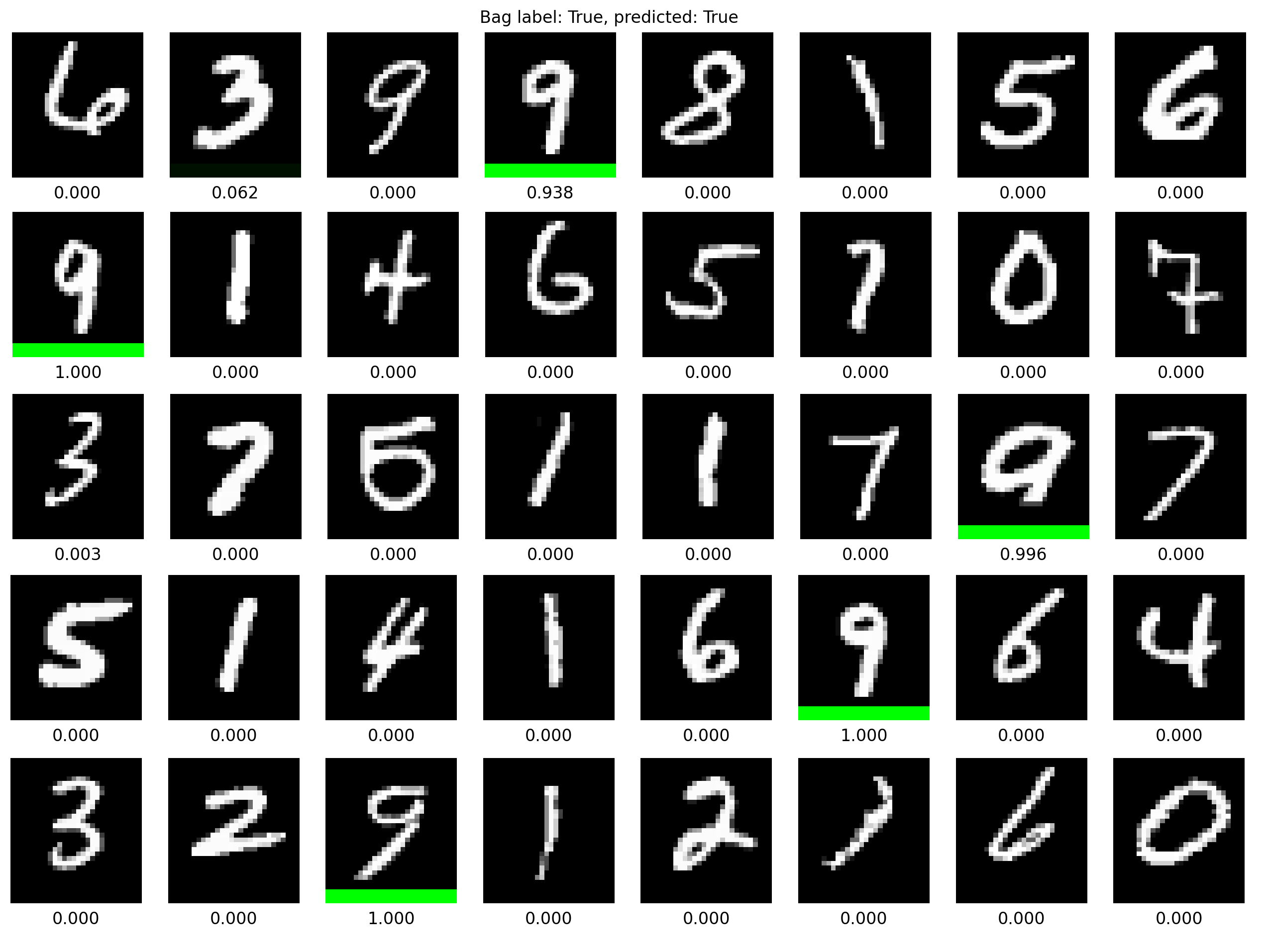}%
\caption{Examples of testing bags from MNIST-MIL-2, which belong to the true
positive subset}%
\label{f:tp01_mnist2}%
\end{center}
\end{figure}

The first subset of five randomly selected testing bags is shown in Fig.
\ref{f:tp01_mnist2}. Every bag belongs to class 1 (true) because it has
neighboring digits \textquotedblleft9\textquotedblright\ and \textquotedblleft%
3\textquotedblright. The algorithm successfully recognizes bags as elements of
class 1 because it finds digit \textquotedblleft9\textquotedblright\ with
neighbor \textquotedblleft3\textquotedblright\ in every bag. The importance
$v_{i}$ of patches (digits) are shown under every patch image. It can be seen
from Fig. \ref{f:tp01_mnist2} that digit \textquotedblleft9\textquotedblright%
\ has the largest value of $v_{i}$. At that, all classified digits
\textquotedblleft9\textquotedblright\ do not have neighbor digits
\textquotedblleft3\textquotedblright.%

\begin{figure}
[ptb]
\begin{center}
\includegraphics[
height=2.8072in,
width=3.7247in
]%
{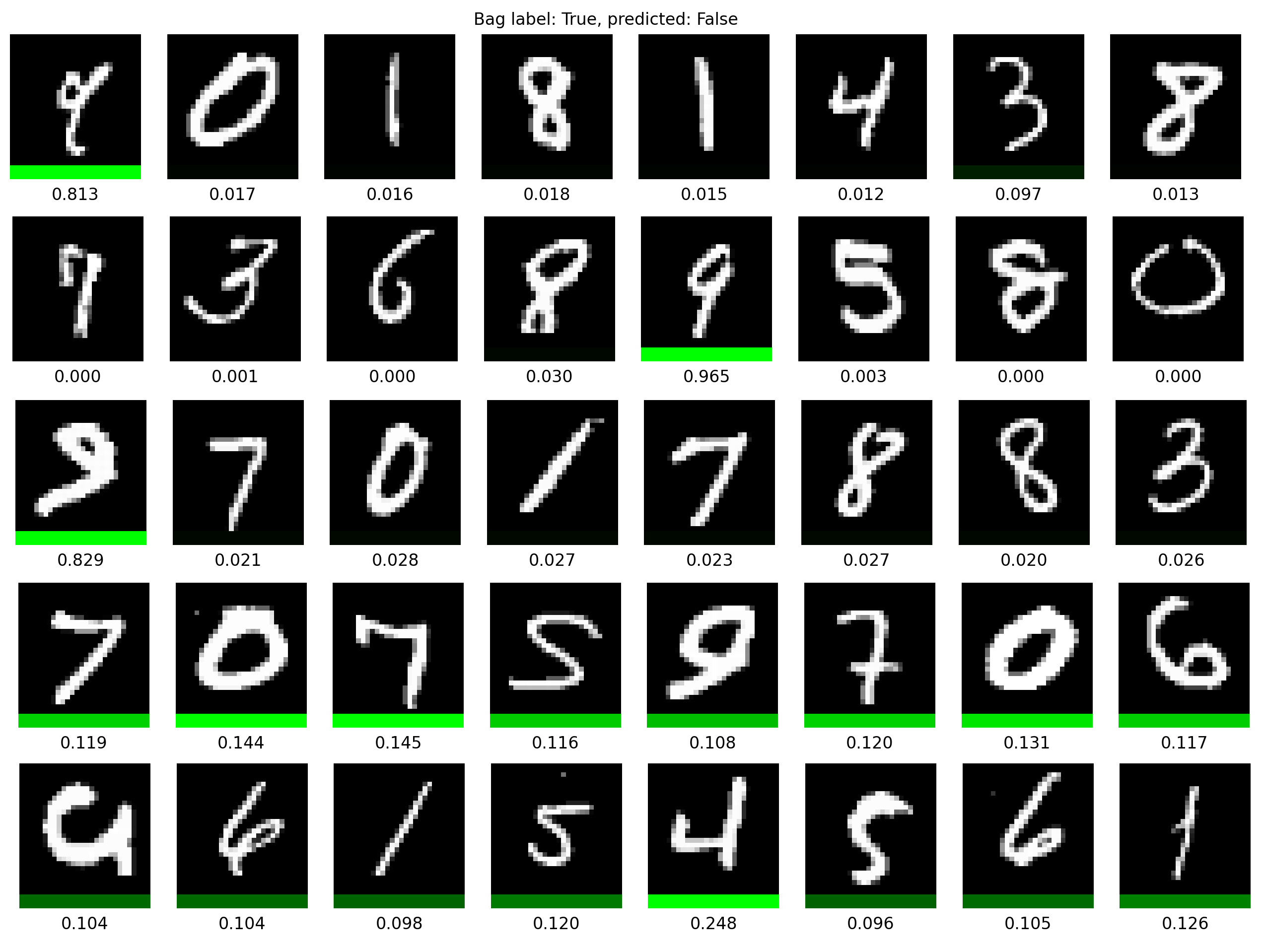}%
\caption{Examples of testing bags from MNIST-MIL-2, which belong to the true
negative subset}%
\label{f:tn01_mnist2}%
\end{center}
\end{figure}

Fig. \ref{f:tn01_mnist2} illustrates another case when examples of testing
bags, which belong to the true negative subset, are shown. One can see from
\ref{f:tn01_mnist2} that every bag contains digits \textquotedblleft%
9\textquotedblright\ without adjacent digits \textquotedblleft%
3\textquotedblright, i.e., the bags belong to class 1. However, the classifier
incorrectly recognizes digits \textquotedblleft9\textquotedblright.%

\begin{figure}
[ptb]
\begin{center}
\includegraphics[
height=2.762in,
width=3.679in
]%
{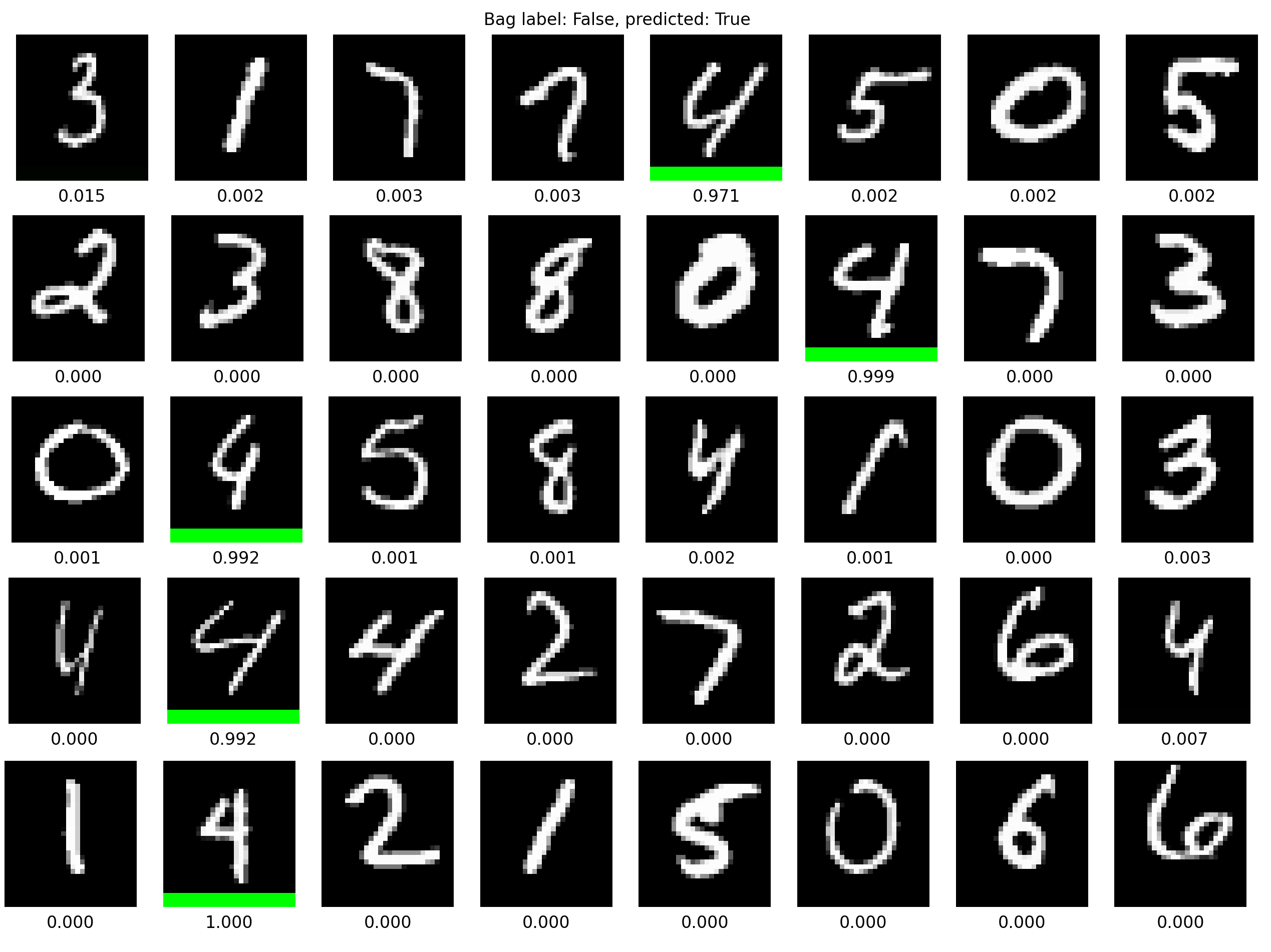}%
\caption{Examples of testing bags from MNIST-MIL-2, which belong to the false
positive subset}%
\label{f:fp01_mnist2}%
\end{center}
\end{figure}

The false positive examples, when bags belong to class 0 because there are no
digits \textquotedblleft9\textquotedblright\ in the bags, are depicted in Fig.
\ref{f:fp01_mnist2}. The classifier incorrectly recognizes digits
\textquotedblleft4\textquotedblright\ as \textquotedblleft9\textquotedblright%
\ which do not have adjacent digits \textquotedblleft3\textquotedblright.%

\begin{figure}
[ptb]
\begin{center}
\includegraphics[
height=2.7821in,
width=3.6815in
]%
{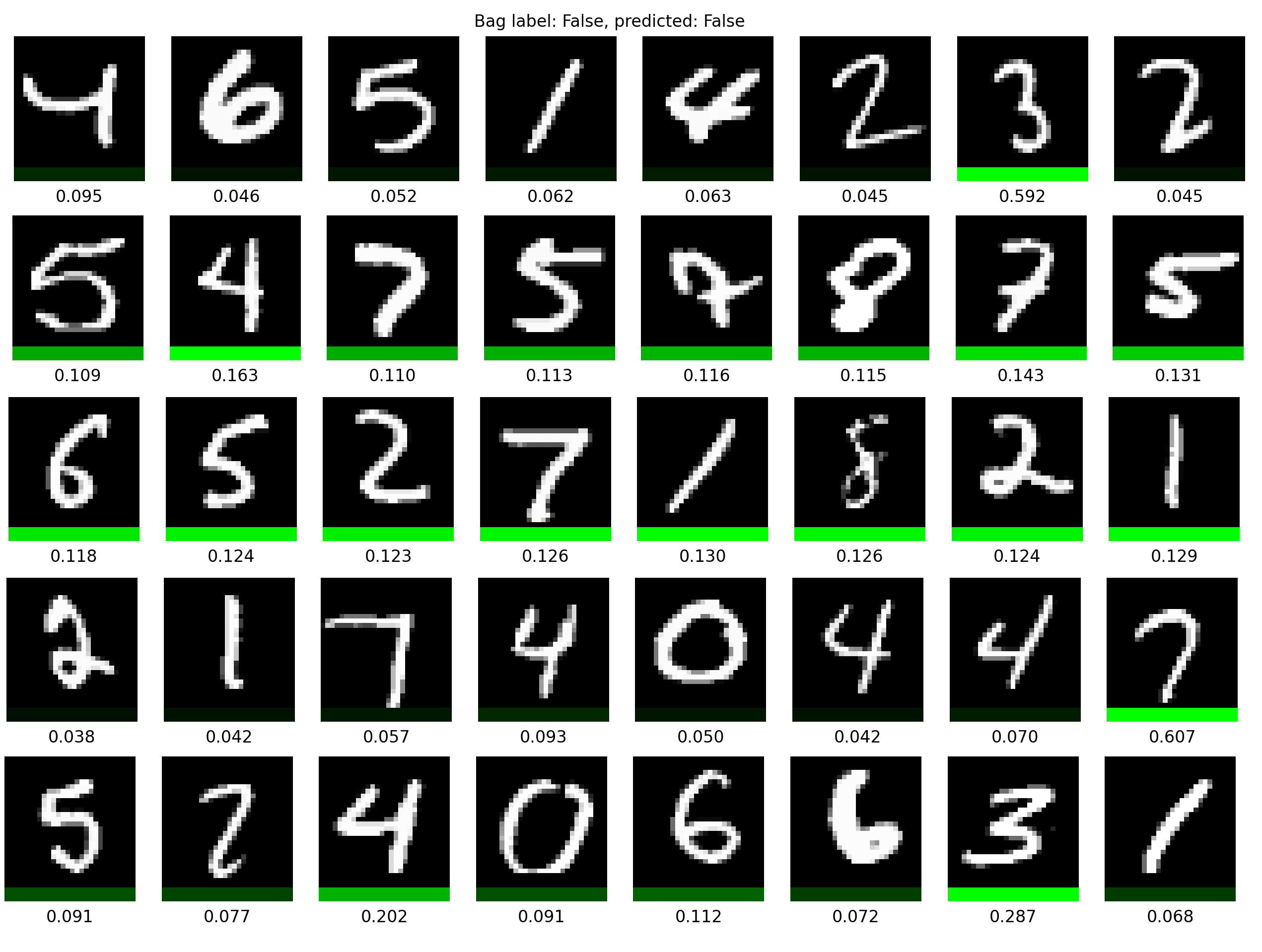}%
\caption{Examples of testing bags from MNIST-MIL-2, which belong to the true
negative subset}%
\label{f:fn01_mnist2}%
\end{center}
\end{figure}

Examples from the true negative subset of MNIST-MIL-2 are shown in Fig.
\ref{f:fn01_mnist2}. Bags do not have digits \textquotedblleft%
9\textquotedblright\ and the classifier does not find these digits.%

\begin{figure}
[ptb]
\begin{center}
\includegraphics[
height=2.8552in,
width=3.8183in
]%
{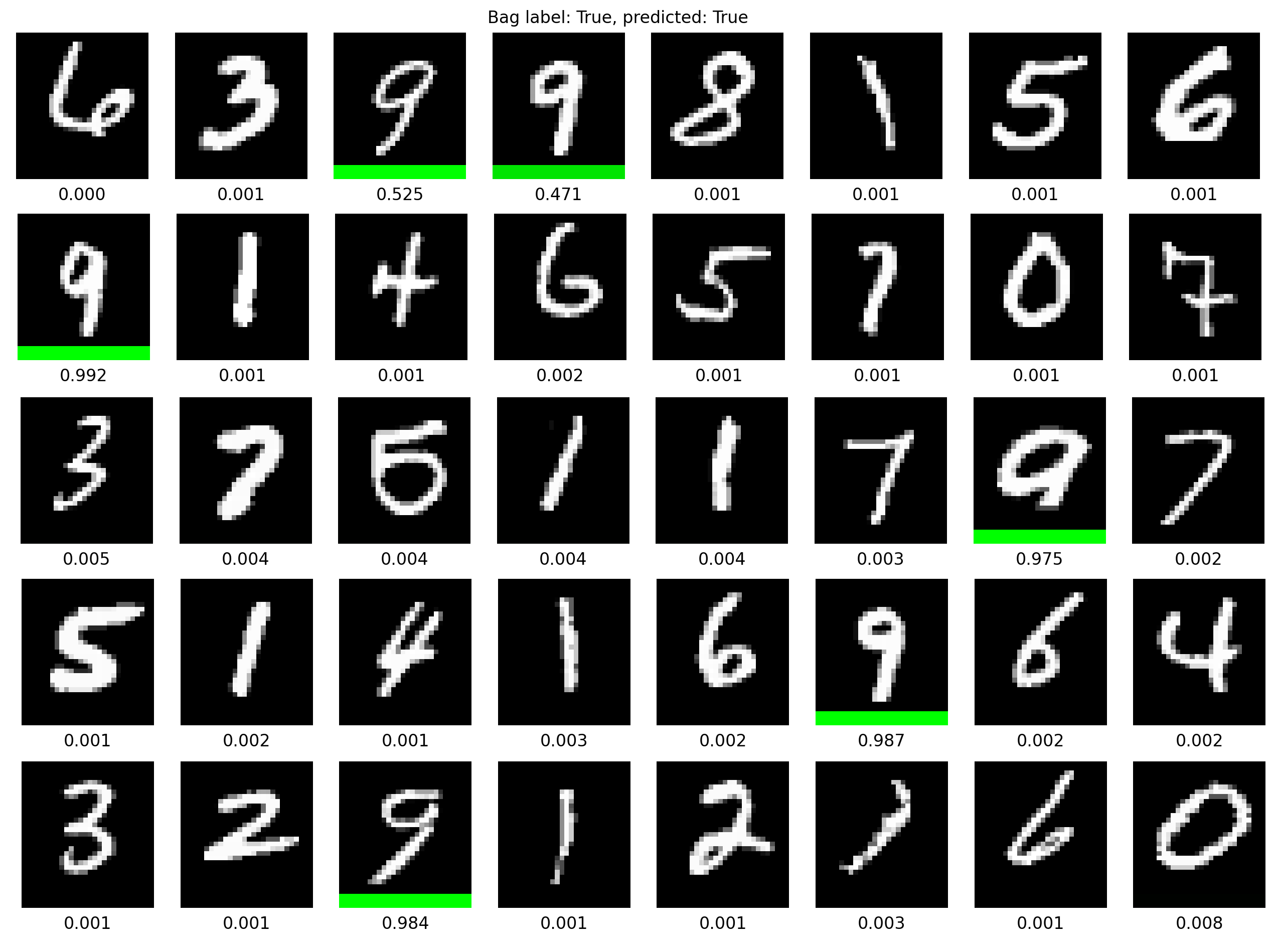}%
\caption{Examples of testing bags from MNIST-MIL, which belong to the true
positive subset}%
\label{f:tp01_mnist0}%
\end{center}
\end{figure}

\begin{figure}
[ptb]
\begin{center}
\includegraphics[
height=2.8867in,
width=3.7948in
]%
{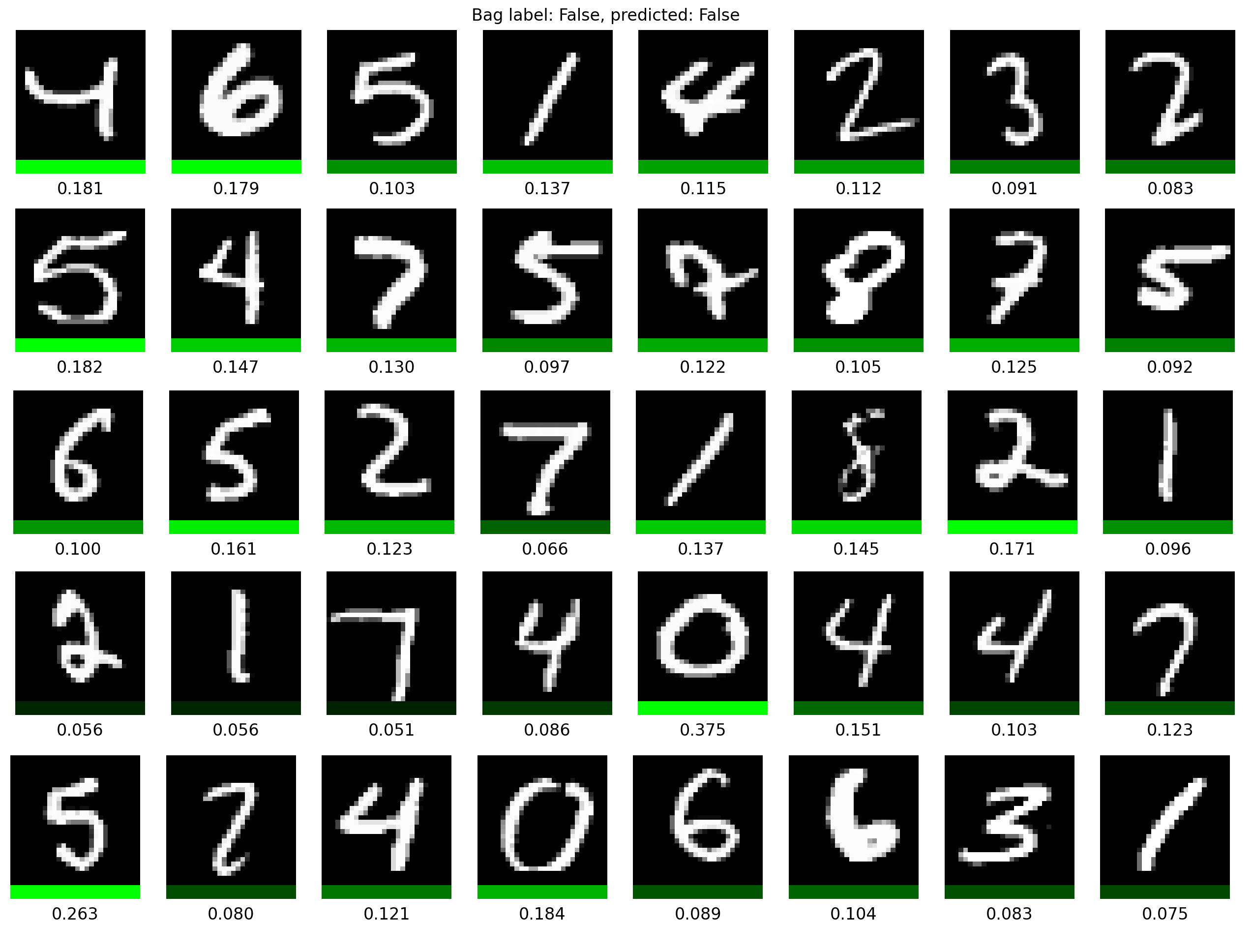}%
\caption{Examples of testing bags from MNIST-MIL, which belong to the false
negative subset}%
\label{f:tn01_mnist0}%
\end{center}
\end{figure}

\begin{figure}
[ptb]
\begin{center}
\includegraphics[
height=2.8878in,
width=3.8089in
]%
{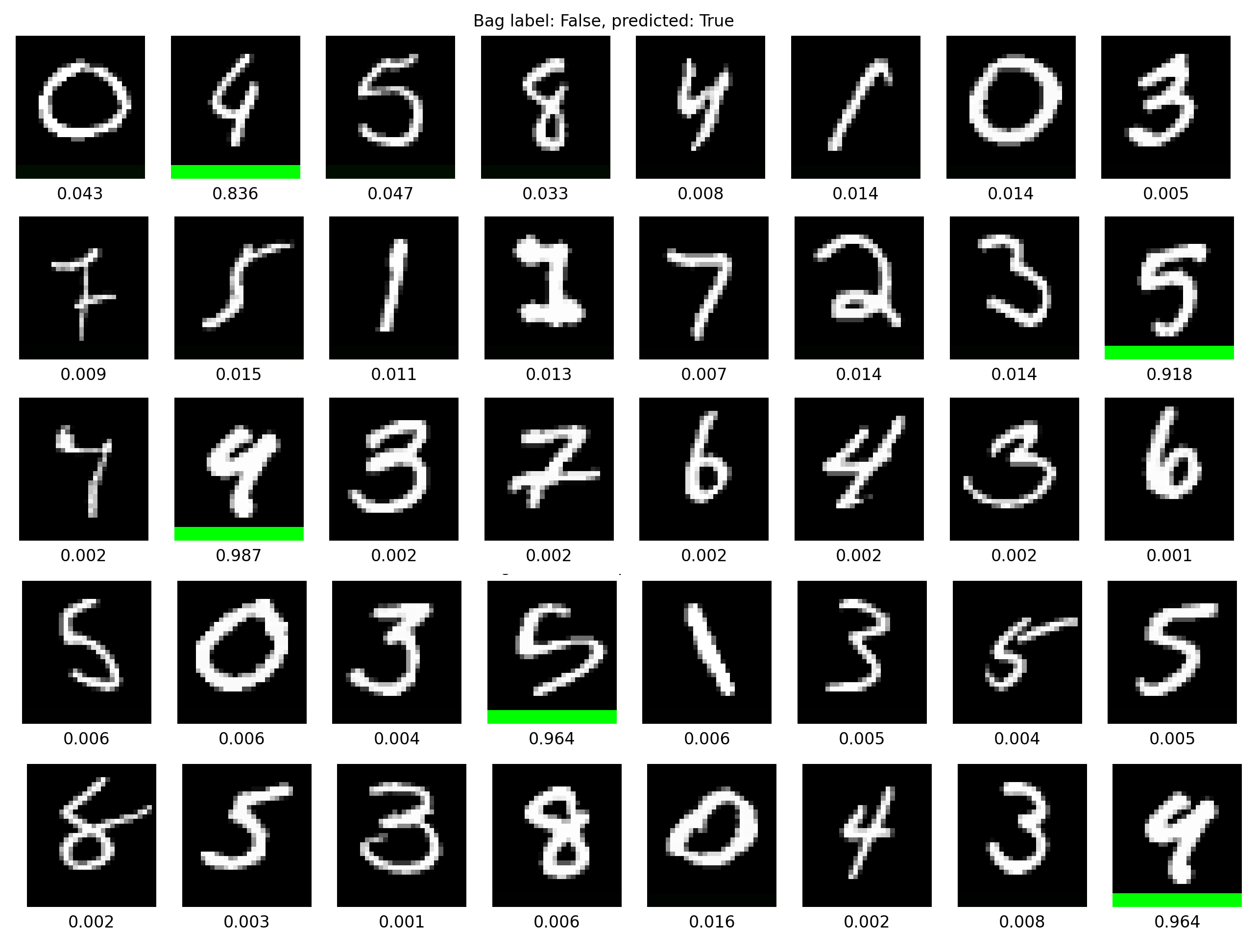}%
\caption{Examples of testing bags from MNIST-MIL, which belong to the false
positive subset}%
\label{f:fp01_mnist0}%
\end{center}
\end{figure}

\begin{figure}
[ptb]
\begin{center}
\includegraphics[
height=2.8643in,
width=3.7948in
]%
{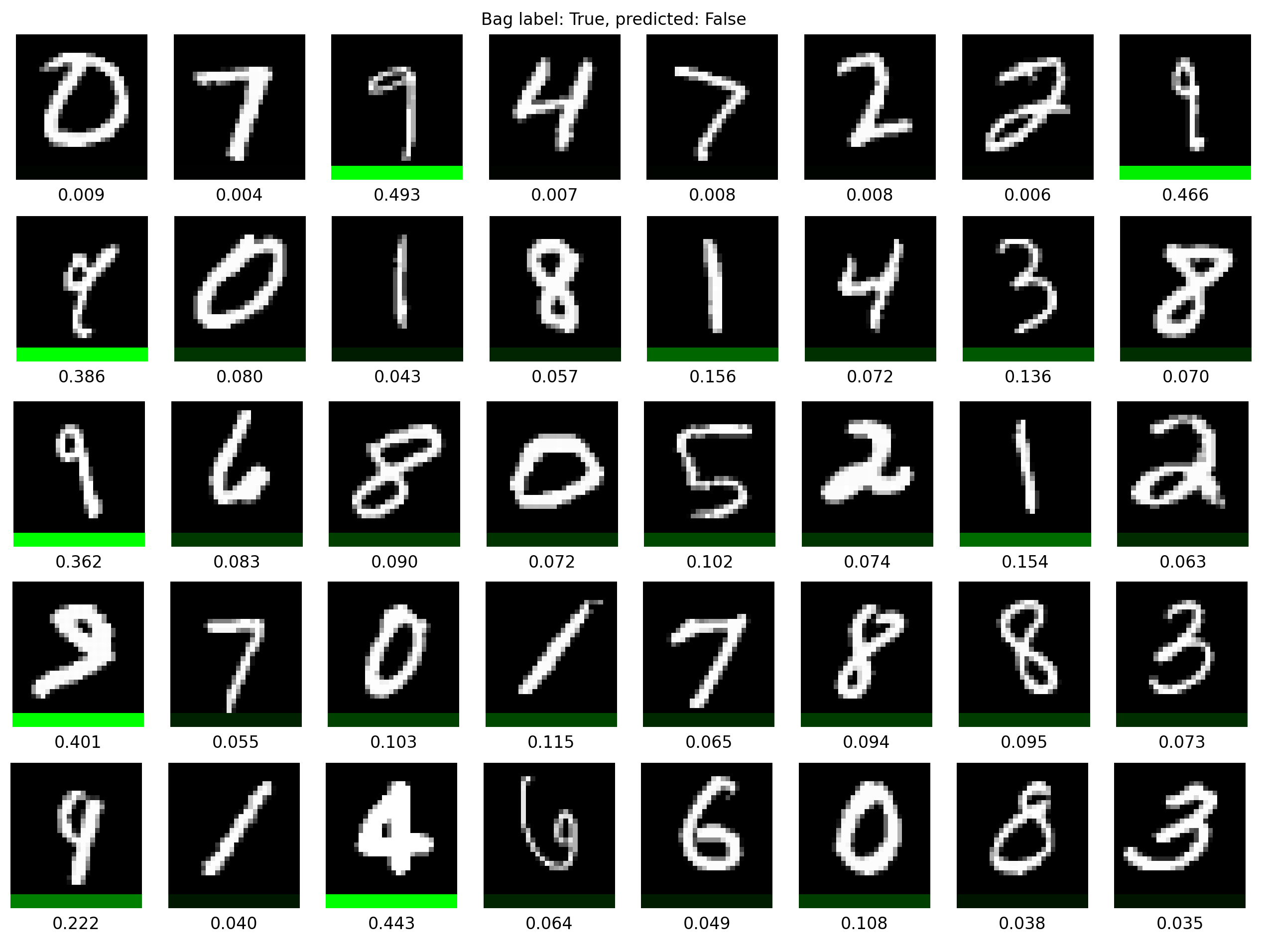}%
\caption{Examples of testing bags from MNIST-MIL, which belong to the true
negative subset}%
\label{f:fn01_mnist0}%
\end{center}
\end{figure}

Similar examples of bags from MNIST-MIL dataset are shown in Figs.
\ref{f:tp01_mnist0}-\ref{f:fn01_mnist0}. Classes of bags are defined by the
simplest condition when digit \textquotedblleft9\textquotedblright\ is
available in the bag.%

%

%

%

The classification accuracy and F1 measure of the proposed model MAMIL and the
attention MIL \cite{Ilse-etal-18} are given in Table \ref{t:MNIST_results}. It
can be seen from Table \ref{t:MNIST_results} that MAMIL outperforms AbDMIL for
all dataset. The F1 measure is used for comparison and analyzing the models
because all dataset are imbalanced. It is interesting to conclude from the
numerical results that MAMIL significantly outperforms AbDMIL for datasets
with complex conditions of classes. In particular, the dataset MNIST-MIL-3 has
the most restrictive condition (digit \textquotedblleft9\textquotedblright%
\ without neighbor \textquotedblleft3\textquotedblright\ and digit
\textquotedblleft7\textquotedblright\ without neighbor \textquotedblleft%
4\textquotedblright), and the difference between F1 measures for MAMIL and
AbDMIL, respectively, is $0.312$.%

\begin{table}[tbp] \centering
\caption{The accuracy and F1 measures for models AbDMIL and MAMIL trained on different datasets formed from MNIST}%
\begin{tabular}
[c]{|c|c|c|c|}\hline
Datasets & Method & Accuracy & F1\\\hline
MNIST-MIL-1 & AbDMIL & $0.806$ & $0.581$\\\hline
MNIST-MIL-1 & MAMIL & $0.987$ & $0.945$\\\hline
MNIST-MIL-2 & AbDMIL & $0.909$ & $0.910$\\\hline
MNIST-MIL-2 & MAMIL & $0.980$ & $0.979$\\\hline
MNIST-MIL-3 & AbDMIL & $0.874$ & $0.601$\\\hline
MNIST-MIL-3 & MAMIL & $0.866$ & $0.913$\\\hline
MNIST-MIL & AbDMIL & $0.973$ & $0.976$\\\hline
MNIST-MIL & MAMIL & $0.983$ & $0.985$\\\hline
\end{tabular}
\label{t:MNIST_results}%
\end{table}%

\begin{figure}
[ptb]
\begin{center}
\includegraphics[
height=1.5954in,
width=2.825in
]%
{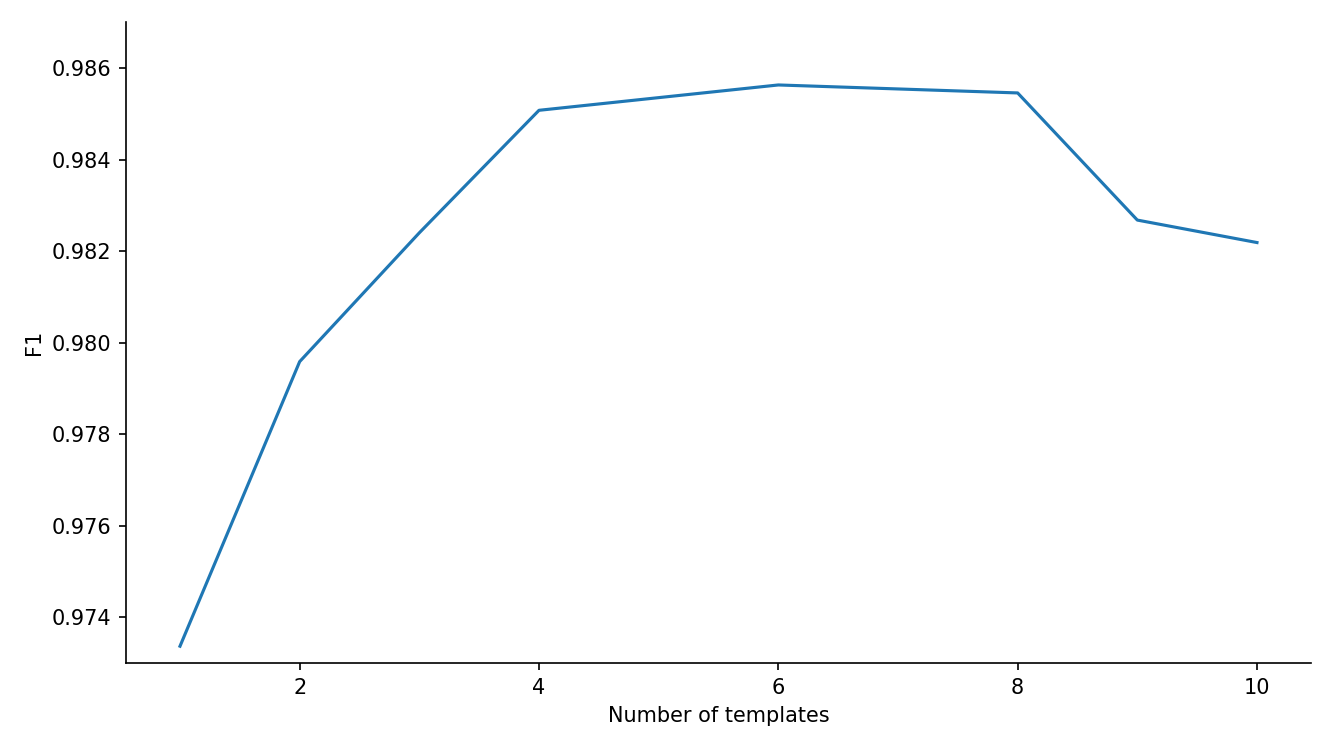}%
\caption{The F1 measure as a function of parameter $C$ (the number of
templates)}%
\label{f:templates}%
\end{center}
\end{figure}

To study how the number of templates impacts on the accuracy measure, we
consider the most interesting case of the MNIST-MIL-3 dataset. Fig.
\ref{f:templates} shows how the F1 measure depends on parameter $C$. It is
interesting to observe that there is some optimal values ($C=6$ for the
considered case) of the template number which provides the largest value of
accuracy. However, the optimal value of $C$ can be found only by its
enumerating in some interval if the number of types of patches is unknown.

\subsection{Datasets Musk1, Musk2, Fox, Tiger, Elephant}

In order to compare MAMIL and the AbDMIL models, we also train the
corresponding models on well-known datasets Musk1, Musk2, Fox, Tiger,
Elephant. It is important to point out that the AbDMIL prediction accuracy
measures are obtained for these datasets in \cite{Ilse-etal-18}. Moreover,
they are compared with many available MIL models, including mi-SVM
\cite{Andrews-etal-02}, MI-SVM \cite{Andrews-etal-02},
MI-Kernel\cite{Gartner-etal-02}, EM-DD \cite{Zhang-Goldman-02}, mi-Graph
\cite{Zhou-Sun-Li-09}, miVLAD \cite{Wei-Wu-Zhou-17}, miFV
\cite{Wei-Wu-Zhou-17}, mi-Net \cite{Wang-Yan-etal-18}, MI-Net
\cite{Wang-Yan-etal-18}, MI-Net with DS \cite{Wang-Yan-etal-18}, MI-Net with
RC \cite{Wang-Yan-etal-18}. The corresponding results can be found in Table 1,
Subsection 4.1 in \cite{Ilse-etal-18}. Therefore, we compare MAMIL only with
the AbDMIL \cite{Ilse-etal-18} to demonstrate its properties. To consider the
AbDMIL method, we take the best accuracy measure among Attention and
Gated-Attention methods.

In our experiments we use the same architecture, optimizer and hyperparameters
of the first Feature Extractor as in the Attention-MIL \cite{Ilse-etal-18}.
Other parameters are the same as in the experiments with MNIST, i.e., $C=10$,
all embeddings $F_{i}$ have the dimensionality which is equal to $128$,
embeddings $T_{i}$, templates $P_{i}$, aggregate embeddings $E_{k}$, vectors
$G$ and $Z$ are of the same size $128$. It should be pointed out that the
whole model is trained without using neighbors because they are not correlated
and may worsen the prediction results. In contrast to the image datasets, the
considered tabular datasets do not have any adjacency relations, i.e.,
neighbors of instances are not defined. However, Table
\ref{t:datasets_results} illustrates that use of multiple attention modules
allows us to significantly enhance the classification accuracy. One can see
from Table \ref{t:datasets_results} that MAMIL outperforms AbDMIL for all
considered datasets.

We also add for comparison purposes the results available for the
Loss-Attention MIL method \cite{shi2020loss} based on the same datasets. which
provide slightly better results in comparison with AbDMIL. It can be seen from
Table \ref{t:datasets_results} that the Loss-Attention method outperforms
MAMIL only for Musk1. However, MAMIL shows better predictions for other datasets.%

\begin{table}[tbp] \centering
\caption{The accuracy measures for AbDMIL and MAMIL trained on datasets Musk1, Musk2, Fox, Tiger, Elephant}%
\begin{tabular}
[c]{|c|c|c|c|}\hline
& AbDMIL & Loss-Attention & MAMIL\\\hline
Musk1 & $0.900$ & $0.917$ & $0.912$\\\hline
Musk2 & $0.863$ & $0.911$ & $0.927$\\\hline
Fox & $0.615$ & $0.712$ & $0.844$\\\hline
Tiger & $0.845$ & $0.897$ & $0.902$\\\hline
Elephant & $0.868$ & $0.900$ & $0.906$\\\hline
\end{tabular}
\label{t:datasets_results}%
\end{table}%

\subsection{Breast Cancer results}

\begin{figure}
[ptb]
\begin{center}
\includegraphics[
height=4.6443in,
width=3.0567in
]%
{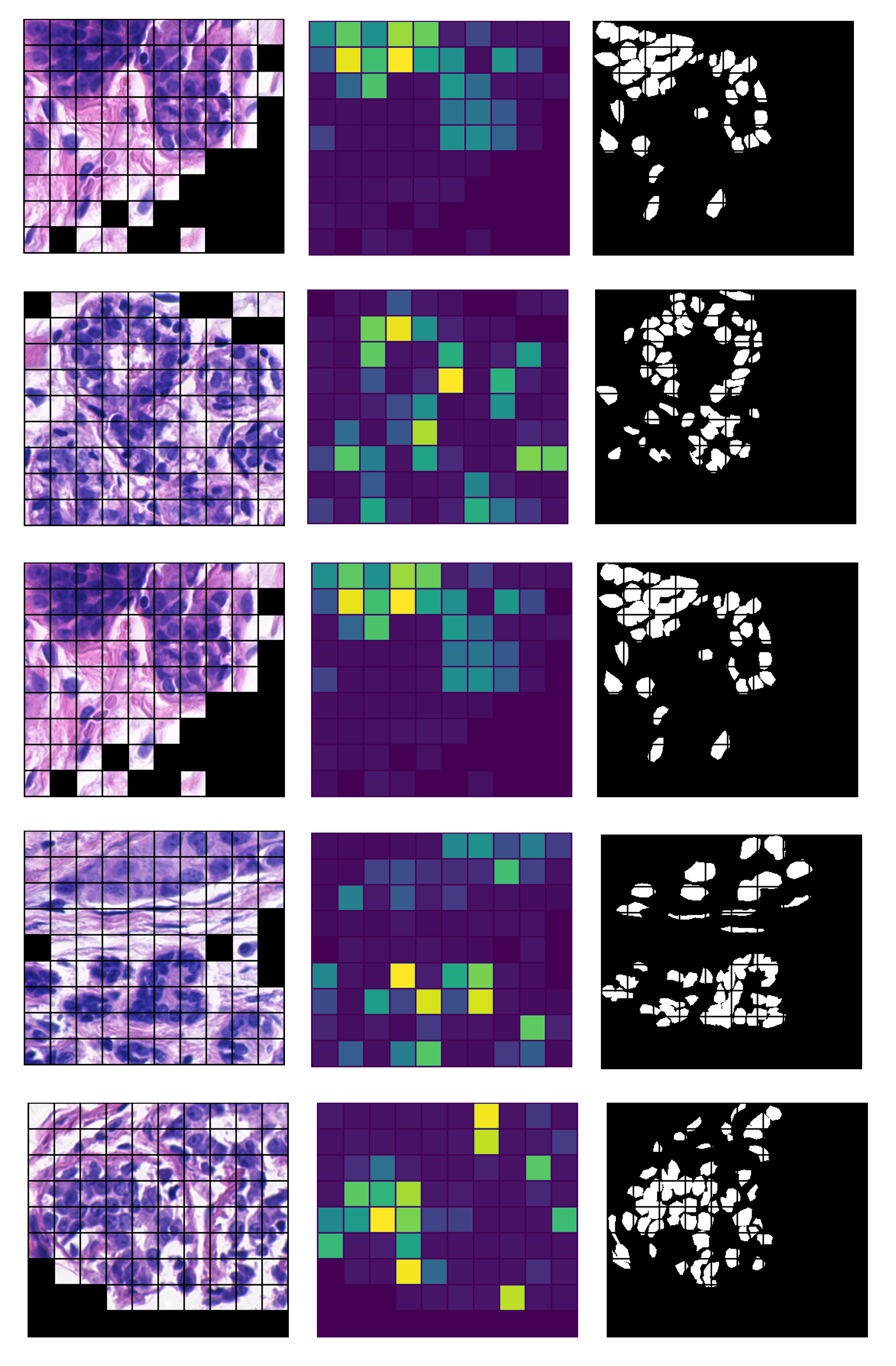}%
\caption{Examples of histology images from the Breast Cancer Cell Segmentation
dataset (the first column), heatmaps constructed in accordance with weights of
patches with images (the second column), ground truth: patches that belong to
class 1 (the third column).}%
\label{f:breast_images}%
\end{center}
\end{figure}

Fig. \ref{f:breast_images} illustrates examples of histology images from the
Breast Cancer Cell Segmentation dataset divided into patches (the first
column), heatmaps created in accordance with weights of patches with images
(the second column), ground truth: patches that belong to class 1 (the third
column). To perform experiments with the Breast Cancer dataset, we use $C=10$
templates for producing aggregate embeddings $E_{k}$. All embeddings $F_{i}$
and $B_{i}$ have the dimensionality which is equal to $128$. Embeddings
$T_{i}$, templates $P_{i}$, aggregate embeddings $E_{k}$, vectors $G$ and $Z$
are of size $256$. Heatmaps are computed by using (\ref{Atten_MIL_61}%
)-(\ref{Atten_MIL_62}). One can see from Fig. \ref{f:breast_images} that there
is a matching between the heatmaps and the ground truth. Moreover, the bag
classification accuracy for the training regime is $\allowbreak\allowbreak
0.968$ whereas for the testing regime is $0.917$.%

\section{Conclusion}

A new model for solving the MIL problem by using the multiple attention
mechanism has been proposed. It differs from many models by several
peculiarities. First, it takes into account neighbors of each instance from a
bag. Second, it uses a specific scheme in the form of the neighborhood
attention for joining the instance and its neighbor information. Third, it
uses a set of attention operations to join information from all instances and
its neighbors in the bag. Fourth, it directly explains results of classification.

Numerical experiments have demonstrated that MAMIL is comparable with the
Loss-Attention model \cite{shi2020loss} and the AbDMIL model
\cite{Ilse-etal-18} and outperforms for most datasets analyzed. In contrast to
these models, MAMIL is more flexible because a proper choice of numbers of
templates and neighbors may significantly improve the model.

MAMIL is not only flexible, but it is rather general and has several elements
which can be changed. First, the concatenation operation is used to unite
embeddings of a patch and its neighbors. However, this operation can be
replaced with the weighted sum of the corresponding embedding vectors like the
attention scheme. Moreover, parameters of the attention can be trainable. This
replacement may compensate anomalous \textquotedblleft mean\textquotedblright%
\ neighborhood embeddings and reduce the dimensionality of joint embeddings.
Another interesting direction for the model modification is extend the
proposed scheme to the neighboring bags. The idea is that neighboring bags
from a dataset like neighboring instances may also improve the classification
properties. In other words, the proposed scheme of multiple attention modules
can be extended to the level of bags. Moreover, it is interesting to consider
a case of unlabeled bags, i.e., to consider the self-supervised learning model
with clusterization. The above ideas can be regarded as interesting directions
for further research.

\section*{Acknowledgement}

This work is supported by the Russian Science Foundation under grant 21-11-00116.

\bibliographystyle{unsrt}
\bibliography{Boosting,Expl_Attention,Explain,Explain_med,Lasso,MIL,MYBIB,Robots}

\end{document}